\documentclass{article}

\title{An Analysis of Arithmetic Constraints on Integer Intervals}
 
\author{Krzysztof R. Apt \\
  {\small CWI, P.O. Box 94079, 1090 GB Amsterdam, the Netherlands and} \\
  {\small University of Amsterdam, the Netherlands}
  \\
  Peter Zoeteweij\footnote { The work of the second author was
    supported by NWO, The Netherlands Organization for Scientific
    Research, under project number 612.069.003.
    } \\
  {\small Faculty of Electrical Engineering, Mathematics and Computer
    Science, } \\
{\small Delft University of Technology, P.O. Box 5031, 2600 GA
    Delft, the Netherlands} } 

\usepackage{url}
\usepackage{amsmath}
\usepackage{latexsym}
\usepackage{alltt}
\usepackage{xspace}

\newcommand{\Proof}{\NI
                    {\bf Proof.}\ }
\newtheorem{theorem}{Theorem}[section]
\newtheorem{lemma}[theorem]{Lemma}
\newtheorem{note}[theorem]{Note}
\newtheorem{exa}[theorem]{Example}
\newenvironment{example}{\begin{exa} \rm}{\end{exa}}
\newtheorem{defined}[theorem]{Definition}
\newenvironment{definition}{\begin{defined} \rm}{\end{defined}}

\newcommand{\eclipse}{ECL$^i$PS$^e$\xspace}
\newcommand{\opint}{\operatorname{int}}
\newcommand{\ophull}{\operatorname{hull}}
\newcommand{\opmin}{\operatorname{min}}
\newcommand{\opmax}{\operatorname{max}}
\newcommand{\opsign}{\operatorname{sign}}

\newcommand{\appd}{$\mathbf{d}$\xspace}
\newcommand{\appdu}{$\mathbf{d_u}$\xspace}
\newcommand{\appdo}{$\mathbf{d_o}$\xspace}
\newcommand{\appp}{$\mathbf{p}$\xspace}
\newcommand{\apppu}{$\mathbf{p_u}$\xspace}
\newcommand{\apppo}{$\mathbf{p_o}$\xspace}
\newcommand{\appf}{$\mathbf{f}$\xspace}
\newcommand{\appfm}{$\mathbf{f_m}$\xspace}
\newcommand{\appfs}{$\mathbf{f_s}$\xspace}
\newcommand{\appfe}{$\mathbf{f_e}$\xspace}

\newcommand{\LL}{\mbox{$\ldots$}}
\newcommand{\oldbfe}[1]{\begin{bfseries}\emph{#1}\end{bfseries}}
\newcommand{\p}[2]{\langle #1 \ ; \ #2 \rangle}
\newcommand{\C}[1]{\mbox{$\{{#1}\}$}}
\newcommand{\ES}{\mbox{$\emptyset$}}
\newcommand{\te}{\mbox{$\exists$}}
\newcommand{\fa}{\mbox{$\forall$}}
\newcommand{\HB}{\hfill{$\Box$}}
\newcommand{\NI}{\noindent}
\newcommand{\floor}[1]{\lfloor #1 \rfloor}
\newcommand{\ceiling}[1]{\lceil #1 \rceil}
\newcommand{\adjceiling}[1]{\left \lceil #1 \right \rceil}
\newcommand{\adjfloor}[1]{\left \lfloor #1 \right \rfloor}
\newcommand{\sse}{\mbox{$\:\subseteq\:$}}

\begin{document}

\date{}
\maketitle

\begin{abstract}
  Arithmetic constraints on integer intervals are supported in many
  constraint programming systems.  We study here a number of
  approaches to implement constraint propagation for these
  constraints.  To describe them we introduce integer interval
  arithmetic.  Each approach is explained using appropriate proof
  rules that reduce the variable domains. We compare these approaches
  using a set of benchmarks.  For the most promising approach we
  provide results that characterize the effect of constraint
  propagation.
\end{abstract}

\section{Introduction}
\label{sec:intro}

\subsection{Motivation}
\label{subsec:motivation}

The subject of arithmetic constraints on reals has attracted a
great deal of attention in the literature. In contrast, arithmetic
constraints on integer intervals have not been studied even though
they are supported in a number of constraint programming systems. In
fact, constraint propagation for them is present in \eclipse{},
SICStus Prolog, GNU Prolog, ILOG Solver and undoubtedly most of the
systems that support constraint propagation for linear constraints on
integer intervals.  Yet, in contrast to the case of linear
constraints --- see notably \cite{HS03} ---
 we did not encounter in the literature any analysis of
this form of constraint propagation.  

In this paper we study these constraints in a systematic
way.  It turns out that in contrast to linear constraints on integer
intervals there are a number of natural approaches to
constraint propagation for these constraints. 
They differ in the extent to which the constraints are decomposed.

Even though arithmetic constraints on integer intervals
need not be decomposed into atomic arithmetic constraints, as is common 
practice for constraints on reals, we found that it is
beneficial to do so: it allows for efficient scheduling of the reduction 
rules and for reuse of auxiliary variables for common subterms between 
constraints.

It could be argued that since integer arithmetic is a special case of
real arithmetic, specialized constraint propagation methods for
integer arithmetic constraints are not needed. Indeed, a constraint
satisfaction problem (CSP) involving arithmetic constraints on integer
variables can be solved using any known method for constraints on
reals, with additional constraints ensuring that the variables assume
only integer values.
This was suggested
in~\cite{BenOld97} and is implemented, for example, in
RealPaver \cite{realpaver}.
However, a dedicated study and implementation of
the integer case is beneficial for a number of reasons.
\begin{itemize}

\item In some cases the knowledge that we are dealing with integers
  yields a stronger constraint propagation than the approach through
  the constraint propagation for arithmetic constraints on reals.
This can be also beneficial 
when we are dealing with hybrid problems that involve
arithmetic constraints on both integer and real variables.
   \item The `indirect' approach through the reals is based on
      floating-point numbers, which are of limited precision.
      This implies that no exact representation exists for integers
      outside certain bounds.  We believe that it should be possible
      to deal with large integers precisely, and that we should not
      revert to a floating-point representation when other options
      exist.  Using a library
      like GNU MP \cite{gmp} we can use
      arbitrary length integers (called multiple \emph{precision} integers
      in GNU MP), whose size is limited only by the
      available memory.

   \item Since arithmetic constraints on integer intervals are supported
      in a number of constraint programming systems, it is natural to
      investigate in a systematic way various approaches to their
      implementation. The approaches based on the integers are
      amenable to a clear theoretical analysis.  In particular, in Section
      \ref{sec:characterization} and Subsection \ref{subsec: weak} we
      provide the characterization results that clarify the effect of
      constraint propagation for the approach that emerged in our studies as
      the fastest.

\end{itemize}
An example that supports the first argument is the constraint $x\cdot y = z$,
where $-3\leq x \leq 3$, $-1 \leq y \leq 1$, and $1\leq z \leq 2$.
When all variables are integers, there are no solutions having $x=3$ or $x=-3$,
and the constraint propagation methods that we consider here will
actually remove these values from the domain of $x$. However, if these
variables are considered to be reals, these values may not be removed,
and solving the integer problem through constraint propagation methods for
constraints on reals may lead to a larger search space.

As an indication that integer representation is not entirely a theoretical
issue, consider the following benchmark from \cite{BenOld97}.
Find $n$ integers $x_1,\ldots,x_n$, $1\leq x_i \leq n$, verifying the
conditions
\[
\sum_{i=1}^{n}x_i = \sum_{i=1}^{n}i,\quad
\prod_{i=1}^{n}x_i = \prod_{i=1}^{n}i,\quad
x_1\leq x_2 \leq \ldots \leq x_n.
\]
For $n=10$ the initial maximum value of the left-hand side expression of the
second constraint equals $10^{10}$, which exceeds $2^{32}$, the number of
values that can be represented as 32-bit integers. For $n=16$,
there is already no signed integer representation
of this bound in 64 bits.

To show that arbitrary length integers can be affordable,
Table~\ref{tab:nlbench} shows timing results for three small C++ programs
that solve the above benchmark via a basic branch-and-propagate search process.
These programs differ only in the representation of the bounds of the variables,
and in the signature of the arithmetic operations applied to these bounds:
64-bit floating point numbers (\texttt{double}), 64-bit integers
(\texttt{long} \texttt{long}), and arbitrary length integers
(using the \texttt{mpz} data type of the GNU MP library).
The programs were compiled using the same optimization flags as the default
build of the GNU MP library, and the reported CPU times are user
time in seconds, measured by the \texttt{time} command on
a 1200 MHz AMD Athlon CPU.

The results for
64-bit integers and $n\geq 16$ could be computed by initializing the upper
bound of the auxiliary variable equated to the product of all problem
variables to $n!$, which works for $n<20$.
These results indicate that on our hardware, the 64-bit integer and
floating-point implementations are equally efficient, while for these specific
problem instances,
the cost of using arbitrary length integers is roughly a factor four.
Note that in a full-fledged constraint solver, this overhead would be far
less prominent, because compared to these small C++ programs, a large part
of the execution time is spent in the framework that coordinates the
computation (cf. the results in Subsection~\ref{subsec:results}).
\begin{table}
\begin{centering}
\begin{tabular}{|r|r|rrr|}
\hline
\multicolumn{2}{|c}{}&\multicolumn{3}{c|}{CPU time (sec.)}\\
\multicolumn{1}{|c}{$n$}  & \multicolumn{1}{r}{solutions} & \texttt{double}&\texttt{long} \texttt{long}  &  \texttt{mpz}\\
\hline
13 &   22 &   0.44 &   0.41 &   1.69 \\
14 &   60 &   1.37 &   1.35 &   5.27 \\
15 &  159 &   4.45 &   4.50 &  17.44 \\
16 &  377 &  14.54 &  15.04 &  57.31 \\
17 &  377 &  32.66 &  33.54 & 128.26 \\
18 & 1007 & 106.77 & 110.98 & 419.74 \\
\hline
\end{tabular}\\
\end{centering}
\caption{Comparison of timing results for various representations}
\label{tab:nlbench}
\end{table}

\subsection{Plan of the Paper}

In the next section we provide the relevant background material on
CSPs and arithmetic constraints.  The
unifying tool in our analysis is integer interval arithmetic that is
modeled after the real interval arithmetic (see, e.g., \cite{HJvE01}).
There are, however, essential differences since we deal with integers
instead of reals. For example, the product of two integer intervals
does not need to be an integer interval.  In Section
\ref{sec:interval} we introduce integer interval arithmetic and
establish the basic results.  Then in Section \ref{sec:intermezzo} we
show that using integer interval arithmetic we can define succinctly
the well-known constraint propagation for linear constraints on
integer intervals.

The next three sections, \ref{sec:first}, \ref{sec:second} and
\ref{sec:third}, form the main part of the paper. We introduce there
three approaches to constraint propagation for arithmetic constraints
on integer intervals.
They differ in the way the constraints are treated: either they are left intact, or
the multiple occurrences of variables are eliminated, or the constraints 
are decomposed into a set of atomic constraints.
In Section~\ref{sec:box} we discuss how these three approaches relate to
various methods used to solve arithmetic constraints on reals.

Then in Section \ref{sec:characterization} we characterize 
the effect of constraint propagation for the atomic constraints.
In Section \ref{sec-experiments} we discuss in detail our
implementation of the alternative approaches, and in
Section~\ref{sec:experiments} we describe the experiments that were performed
to compare them.
They indicate that decomposition of the constraints, combined with
a scheduling of the reduction rules that respects the hierarchical dependencies
between the atomic constraints is superior to the other approaches.
Finally, in Section~\ref{sec:conclusions} we provide the conclusions.

The preliminary results of this work were reported in 
\cite{Apt03} and \cite{AZ04}.

\section{Preliminaries}
\label{sec:preliminaries}

\subsection{Constraint Satisfaction Problems}

We now review the standard concepts of a constraint and of a
constraint satisfaction problem.  Consider a sequence of variables $X
:= x_1, \LL, x_n$ where $n \geq 0$, with respective domains $D_1, \LL,
D_n$ associated with them.  So each variable $x_i$ ranges over the
domain $D_i$.  By a \oldbfe{constraint} $C$ on $X$ we mean a subset of
$D_1 \times \LL \times D_n$.  Given an element $d := d_1, \LL, d_n$ of
$D_1 \times \LL \times D_n$ and a subsequence $Y := x_{i_1}, \LL,
x_{i_l}$ of $X$ we denote by $d[Y]$ the sequence $d_{i_1}, \LL,
d_{i_{l}}$. In particular, for a variable $x_i$ from $X$, $d[x_i]$
denotes $d_i$.

A \oldbfe{constraint satisfaction problem}, in short CSP, consists of
a finite sequence of variables $X$ with respective domains ${\cal
  D}$, together with a finite set $\cal C$ of constraints, each on a
subsequence of $X$. We write it as $\p{{\cal C}}{x_1 \in D_1,
  \LL, x_n \in D_n}$, where $X := x_1, \LL, x_n$ and ${\cal D} :=
D_1, \LL, D_n$.

By a \oldbfe{solution} to $\p{{\cal C}}{x_1 \in D_1, \LL, x_n \in D_n}$
we mean an element $d \in D_1 \times \LL \times D_n$ such that for
each constraint $C \in {\cal C}$ on a sequence of variables $X$ we
have $d[X] \in C$.  We call a CSP \oldbfe{consistent} if it has a
solution and \oldbfe{inconsistent} if it does not.
Two CSPs with the same sequence of variables are called
\oldbfe{equivalent\/} if they have the same set of solutions.  In what
follows we consider CSPs whose constraints are defined in a
simple language and when reasoning about them we identify the 
syntactic description of a constraint
with its meaning being the set of tuples that satisfy it.

We view \oldbfe{constraint propagation} as a process of transforming CSPs that maintains
their equivalence. In what follows we define this process by means
of proof rules that operate on CSPs and preserve equivalence.
An interested reader can consult \cite{Apt98a} or \cite{Apt03}
for a precise explanation of 
this approach to describing constraint propagation.

\subsection{Arithmetic Constraints}

To define the arithmetic constraints we use the alphabet that comprises
\begin{itemize}
\item variables,

\item two constants, 0 and 1,

\item the unary minus function symbol `$-$',
  
\item three binary function symbols,  `+',`$-$'and `$\cdot$', all written in the
  infix notation.
\end{itemize}

By an \oldbfe{arithmetic expression} we mean a term formed in this
alphabet and by an \oldbfe{arithmetic constraint} 
a formula of the form
\[
s \: op \: t,
\]
where $s$ and $t$ are arithmetic expressions and $op \in \C{<, \leq, =, \neq, \geq, >}$.
For example
\begin{equation}
x^5 \cdot y^2 \cdot z^4  + 3 x \cdot y^3 \cdot z^5 \leq 10 + 4 x^4 \cdot y^6 \cdot z^2 - y^2 \cdot x^5 \cdot z^4
\label{eq:arithcon}
\end{equation}
is an arithmetic constraint. Here $x^5$ is an abbreviation for $x \cdot x \cdot x \cdot x \cdot x$,
while $3 x \cdot y^3 \cdot z^5$ is an abbreviation for
$x \cdot y^3 \cdot z^5 + x \cdot y^3 \cdot z^5 + x \cdot y^3 \cdot z^5$, 
and similarly
with the other expressions.
If `$\cdot$' is not used in an arithmetic constraint, we call it a \oldbfe{linear constraint}.

By an \oldbfe{extended arithmetic expression} we mean a term formed in the above
alphabet extended by the unary function symbols `$\cdot^n$' and `$\sqrt[n]{\cdot}$' for each $n \geq 1$ and
the binary function symbol `$/$' written in the infix notation.
For example
\begin{equation}
\sqrt[3]{(y^2 \cdot z^4)/(x^2 \cdot u^5)}
\label{eq:extended}
\end{equation}
is an extended arithmetic expression. Here, unlike in (\ref{eq:arithcon}),
$x^5$ is a term obtained by applying the function symbol `$\cdot^5$' to the
variable $x$.
The extended arithmetic expressions will be used only to define constraint
propagation for the arithmetic constraints.

Fix now some arbitrary linear ordering $\prec$ on the variables of the language.
By a \oldbfe{monomial} we mean an integer or a term
of the form
\[
a \cdot x_{1}^{n_1} \cdot \LL   \cdot x_{k}^{n_k}
\]
where $k > 0$, $x_1, \LL, x_k$ are different variables ordered w.r.t.~%
$\prec$, and $a$ is a non-zero integer and $n_1, \LL, n_k$ are
positive integers.  We call then $x_{1}^{n_1} \cdot \LL \cdot
x_{k}^{n_k}$ the \oldbfe{power product} of this monomial.

Next, by a \oldbfe{polynomial} we mean a term of the form
\[
\Sigma^{n}_{i=1} m_i,
\]
where $n > 0$, at most one monomial $m_i$ is an integer, and the power
products of the monomials $m_1, \LL, m_n$ are pairwise different.
Finally, by a \oldbfe{polynomial constraint} we mean an arithmetic
constraint of the form $s \: op \: b$, where $s$ is a polynomial with
no monomial being an integer, $op \in \C{<, \leq, =, \neq, \geq, >}$,
and $b$ is an integer.  It is clear that
by means of appropriate transformation rules we can transform each
arithmetic constraint to a polynomial constraint. For example,
assuming the ordering $x \prec y \prec z$ on the variables, the
arithmetic constraint (\ref{eq:arithcon}) can be transformed to the
polynomial constraint
\[
2x^5 \cdot y^2 \cdot z^4 - 4 x^4 \cdot y^6 \cdot z^2  + 3 x \cdot y^3 \cdot z^5 \leq 10
\]
So, without loss of generality, from now on we shall limit our
attention to polynomial constraints.

Next, let us discuss the domains over which we interpret the arithmetic constraints.
By an \oldbfe{integer interval}, or an \oldbfe{interval} 
in short, we mean
an expression of the form
\[
[a..b]
\]
where $a$ and $b$ are integers;
$[a..b]$ denotes the set of all integers between $a$ and $b$,
including $a$ and $b$.
If $a > b$, we call $[a..b]$ the \oldbfe{empty interval} and denote it by $\ES$.
By a \oldbfe{range} we mean an expression of the form
\[
x \in I
\]
where $x$ is a variable and $I$ is an interval.
Sets of the form $\{x\in\mathcal{Z}|x\geq a\}$ and $\{x\in\mathcal{Z}|x\leq b\}$
are called \oldbfe{extended intervals}.

We link the arithmetic constraints with the notion of a constraint defined in the
previous section by associating in the standard way
with each arithmetic constraint its \oldbfe{interpretation}.
For an arithmetic constraint on
variables $x_1, \LL, x_n$ with respective integer interval domains $D_1, \LL,
D_n$ this is a subset of $D_1 \times \LL \times D_n$.

\section{Integer Set Arithmetic}
\label{sec:interval}

To reason about the arithmetic constraints we employ a generalization of the
arithmetic operations to the sets of integers.
Here and elsewhere ${\cal Z}$, ${\cal N}$, and ${\cal R}$ denote the sets of
all integers, natural numbers, and reals, respectively.

\subsection{Definitions}

For $X,Y$ sets of integers we define the following
operations:

\begin{itemize}

\item addition:
\[
X + Y := \C{x + y \mid x \in X, y \in Y},
\]

\item subtraction:
\[
X - Y := \C{x - y \mid x \in X, y \in Y},
\]

\item multiplication:
\[
X \cdot Y := \C{x \cdot y \mid x \in X, y \in Y},
\]

\item division:
\[
X/Y  := \C{u \in {\cal Z} \mid \te x \in X \te y \in Y \: u \cdot y = x},
\]

\item exponentiation:
\[
X ^ n := \C{x ^n \mid x \in X},
\]
for each natural number $n > 0$,

\item root extraction:
\[
\sqrt[n]{X} := \C{x \in {\cal Z} \mid x ^n \in X},
\]
for each natural number $n > 0$.

\end{itemize}

All the operations except division and root extraction are defined in
the expected way.  We shall return to the division operation
in Section~\ref{sec:third}.  
At the moment it suffices to note the
division operation is defined for all sets of integers, including $Y =
\ES$ and $Y = \C{0}$.  This division operation corresponds to the
following division operation on the sets of reals introduced in
\cite{Rat96}:

\[
X \oslash Y  := \C{u \in {\cal R} \mid \te x \in X \te y \in Y \: u \cdot y = x}.
\]
For an integer or real number $a$ and
$op \in \C{+, -, \cdot, /, \oslash }$ we identify
$a \: op \: X$ with $\C{a} \: op \: X$ and $X \: op \: a$ with $X \: op \: \C{a}$.

To present the rules we are interested in we shall also use the
addition and division operations on the sets of reals.
Addition is defined in the same way as for the sets of integers, and
for division we use the $\oslash$ operator defined above.
In \cite{HJvE01} it is explained
how to implement these operations on, possibly unbounded, real intervals.

Further, given a set $A$ of integers or reals, we define
\[
^{\leq}A := \C{x \in {\cal Z} \mid \te a \in A \: x \leq a},
\]
\[
^{\geq}A := \C{x \in {\cal Z} \mid \te a \in A \: x \geq a},
\]
so for example $^\leq{\mathcal{N}}=\mathcal{Z}$, and
$^\geq\{-1,1\}$ and $^\geq(-2,2)$ both denote the extended interval of all
integers greater than or equal to $-1$, where $(-2,2)$ denotes an
open interval of real numbers.

When limiting our attention to intervals of integers the
following simple observation is of importance.  

\begin{note} \label{note:1}
For $X,Y$ integer intervals and $a$ an integer the following holds:

\begin{itemize}
\item $X \cap Y$, $X + Y, X - Y$ are integer intervals.
\item $X/\C{a}$ is an integer interval.
\item $X \cdot Y$ does not have to be an integer interval, even if $X = \C{a}$ or $Y = \C{a}$.
\item $X/Y$ does not have to be an integer interval. 
\item For each $n >1$, \ $X ^ n$ does not have to be an integer interval.
\item For odd $n >1$, \ $\sqrt[n]{X}$ is an integer interval.
\item For even $n >1$, \ $\sqrt[n]{X}$ is an integer interval or a disjoint union of two integer intervals.
\HB
\end{itemize}
\end{note}

For example in the following cases we get intervals as outcomes:
\[
[2..4] + [3..8] = [5..12],
\]
\[
[3..7] - [1..8] = [-5..6],
\]
\[
\sqrt[3]{[-30..100]} = [-3..4],
\]
\[
\sqrt[2]{[-100..9]} = [-3..3],
\]
while in the following ones not:
\[
[3..3] \cdot [1..2] = \C{3,6},
\]
\[
[3..5]/[-1..2] = \C{-5,-4,-3,2,3,4,5},
\]
\[
[-3..5]/[-1..2] = {\cal Z},
\]
\[
[1..2]^2 = \C{1,4},
\]
\[
\sqrt[2]{[1..9]} = [-3..-1] \cup [1..3].
\]

To deal with the problem that non-interval domains can be produced
by some of the operations we introduce the following operation 
on the sets of integers:
\[
\opint(X) :=    
\left\{ 
\begin{tabular}{ll}
\mbox{smallest integer interval containing $X$} &  \mbox{if $X$ is finite,} \\
${\cal Z}$                                   &  \mbox{otherwise.}
\end{tabular}
\right . 
\]
For example $\opint([3..5]/[-1..2]) = [-5..5]$
and $\opint([-3..5]/[-1..2]) = {\cal Z}$.

\subsection{Implementation}

\label{subsec:implementation}

To define constraint propagation for the arithmetic constraints on integer intervals
we shall use the integer set arithmetic, mainly limited to the integer intervals.
This brings us to the discussion of how to
implement the introduced operations on the integer intervals.
Since we are only interested in maintaining the property that the sets remain integer
intervals or the set of integers ${\cal Z}$
we shall clarify how to implement the
intersection, addition, subtraction and root extraction operations of the integer intervals and the
$\opint(\cdot)$ closure of the multiplication, division and exponentiation operations on the
integer intervals.  The case when one of the intervals is empty is easy to deal with. So we
assume that we deal with non-empty intervals
$[ a..b ]$ and $[ c..d ]$, that is $a \leq b$ and $c \leq d$.

\paragraph{Intersection, addition and subtraction}
It is easy to see that
\[
[ a..b ] \cap [ c..d ] = [ \opmax(a,c)..\opmin(b,d)],
\]
\[
[ a..b ] +[ c..d ] = [a+c\ ..\ b+d],
\]
\[
[ a..b ] - [ c..d ] = [a-d\ ..\ b-c].
\]
So the interval intersection, addition, and subtraction are straightforward to implement.

\paragraph{Root extraction}

The outcome of the root extraction operator applied to an integer
interval will be an integer interval or a disjoint union of two
integer intervals.  We shall explain in Section \ref{sec:first} why it is
advantageous not to apply $\opint(\cdot)$ to the outcome.  This operator can
be implemented by means of the following case analysis.  
\medskip

\NI
\emph{Case 1.} Suppose $n$ is odd. Then
\[
\sqrt[n]{[a..b]} = [\adjceiling{\sqrt[n]{a}} .. \adjfloor{\sqrt[n]{b}}].
\]

\NI
\emph{Case 2.} Suppose $n$ is even and $b < 0$. Then

\[
\sqrt[n]{[a..b]} = \ES.
\]

\NI
\emph{Case 3.} Suppose $n$ is even and $b \geq 0$. Then
\[
\sqrt[n]{[a..b]} = [-\adjfloor{|\sqrt[n]{b}|} .. -\adjceiling{|\sqrt[n]{a_{+}}|}]
              \cup [\adjceiling{|\sqrt[n]{a_{+}}|} .. \adjfloor{|\sqrt[n]{b}|} ]
\]
where $a_{+} := \opmax(0,a)$.

\paragraph{Multiplication}
For the remaining operations we only need to explain how to implement
the $\opint(\cdot)$ closure of the outcome.
First note that \[
\opint([ a..b] \cdot [ c..d])   =  [ \opmin(A)..  \opmax(A) ],  
\]
where $A = \C{a \cdot c, a \cdot d, b \cdot c, b \cdot d}$.  

Using an appropriate case analysis we can actually compute the bounds
of $\opint([ a..b] \cdot [ c..d])$ directly in terms of the bounds of the
constituent intervals.

\paragraph{Division}
In contrast, the $\opint(\cdot)$ closure of the interval division is not so
straightforward to compute. The reason is that, as we shall see in a
moment, we cannot express the result in terms of some simple
operations on the interval bounds.

Consider non-empty integer intervals $[ a..b] $ and $[ c..d ] $.
In analyzing the outcome of $\opint([a..b]/[c..d])$ we distinguish the following
cases.
\medskip

\NI
\emph{Case 1.} Suppose $0 \in [a .. b]$ and $0 \in [c .. d]$.

Then by definition $\opint([a..b]/[c..d]) = {\cal Z}$.
For example, 
\[
\opint([-1..100]/[-2..8]) = {\cal Z}.
\]

\NI
\emph{Case 2.} Suppose $0 \not \in [a .. b]$ and $c = d = 0$.

Then by definition $\opint([a .. b]/[c .. d]) = \ES$.
For example, 
\[
\opint([10..100]/[0..0]) = \ES.
\]

\NI
\emph{Case 3.} Suppose $0 \not \in [a .. b]$ and $c < 0$ and $0 < d$.

It is easy to see that then
\[
\opint([a .. b]/[c .. d]) = [-e .. e],
\] 
where $e = \opmax (|a|, |b|)$.
For example, 
\[
\opint([-100..-10]/[-2..5]) = [-100..100].
\]

\NI
\emph{Case 4.} Suppose $0 \not \in [a .. b]$ and either $c = 0$ and $d \neq 0$ or $c \neq 0$ and $d = 0$.

Then $\opint([a .. b]/[c .. d]) = \opint([a .. b]/([c .. d] -  \C{0}))$.
For example 
\[
\opint([1..100]/[-7..0]) = \opint([1..100]/[-7..-1]).
\]
This allows us to reduce this case to Case 5 below.

\NI
\emph{Case 5.} Suppose $0 \not \in [c .. d]$.

This is the only case when we need to compute $\opint([a .. b]/[c .. d])$
indirectly. First, observe that we have 
\[
\opint([a..b]/[c..d]) \sse [\adjceiling{\opmin(A)} .. \adjfloor{\opmax(A)}],    
\]
where $A = \C{a/c, a/d, b/c, b/d}$.

However, the equality does not need to hold here.  Indeed, note for
example that $\opint([155..161]/[9..11]) = [16..16]$, whereas for $A =
\{155/9, 155/11, 161/9$, $161/11\}$ we have $\adjceiling{\opmin(A)} = 15$
and $\adjfloor{\opmax(A)} = 17$.  The problem is that the value 16 is
obtained by dividing 160 by 10 and none of these two values is an
interval bound.  

This complication can be solved by preprocessing the interval $[c..d]$
so that its bounds are actual divisors of an element of $[a..b]$.
First, we look for the least $c' \in [c..d]$ such
that $\exists x \in [a..b]\ \exists u \in {\cal Z}\ u\cdot c' = x$.
Using a case analysis, the latter property can be established without search.
Suppose for example that $a>0$ and $c>0$. In this case, if 
$c' \cdot \lfloor b/c' \rfloor \geq a$, 
then $c'$ has the required
property. Similarly, we look for the largest 
$d' \in [c..d]$ for which an analogous condition holds. Now
$\opint([a..b] / [c..d]) = [\lceil\opmin(A)\rceil ..
\lfloor\opmax(A)\rfloor]$, where $A = \{a/c', a/d', b/c',
b/d'\}$.

\medskip
In view of this auxiliary computation (in case when $0 \not \in [c..d]$)
we shall introduce in Section \ref{sec-experiments} a modified division operation with a more
direct implementation.

\paragraph{Exponentiation}
The $\opint(\cdot)$ closure of the interval exponentiation is straightforward 
to implement by distinguishing the following cases.
\medskip

\NI
\emph{Case 1.} Suppose $n$ is odd. Then
\[
\opint([a..b]^n) = [a^n ..\: b ^n]. 
\]

\NI
\emph{Case 2.} Suppose $n$ is even and $0 \leq a$. Then
\[
\opint([a..b]^n) = [a^n ..\: b ^n]. 
\]

\NI
\emph{Case 3.} Suppose $n$ is even and $b \leq 0$. Then
\[
\opint([a..b]^n) = [b^n ..\: a^n]. 
\]

\NI
\emph{Case 4.} Suppose $n$ is even and $a < 0$ and $0 < b$. Then
\[
\opint([a..b]^n) = [ 0..  \opmax(a^n, b^n) ].
\]

\subsection{Correctness Lemma}
\label{subsec:correctness}

Given now an extended arithmetic expression $s$ each variable of
which ranges over an integer interval, we define $\opint(s)$ as the
integer interval or the set ${\cal Z}$ obtained by systematically
replacing each function symbol by the application of the $\opint(\cdot)$
operation to the corresponding integer set operation.  For example,
for the extended arithmetic expression $s := \sqrt[3]{(y^2 \cdot
  z^4)/(x^2 \cdot u^5)}$ of (\ref{eq:extended}) we have
\[
\opint(s)= \opint(\sqrt[3]{\opint(\opint(Y^2) \cdot \opint(Z^4))/\opint(\opint(X^2) \cdot \opint(U^5))}),
\]
where we assume that $x$ ranges over $X$, etc.

The discussion in the previous subsection shows how to compute
$\opint(s)$ given an extended arithmetic expression $s$ and the integer
interval domains of its variables.

The following lemma is crucial for our considerations. It is a counterpart of the
so-called `Fundamental Theorem of Interval Arithmetic' established in \cite{Moo66}.
Because we deal here with the integer domains an additional assumption is
needed to establish the desired conclusion.

\begin{lemma}[Correctness] \label{lem:correctness}
Let $s$ be an extended arithmetic expression with the variables
$x_1, \LL, x_n$. Assume that each variable $x_i$ of $s$
ranges over an integer interval $X_i$.
Choose $a_i \in X_i$ for $i \in [1..n]$ and 
denote by $s(a_1, \LL, a_n)$ the result of replacing in $s$
each occurrence of a variable $x_i$ by $a_i$.

Suppose that each subexpression of 
$s(a_1, \LL, a_n)$ evaluates to an integer. Then
the result of evaluating $s(a_1, \LL, a_n)$ is an element of $\opint(s)$.
\end{lemma}
\Proof The proof follows by a straightforward induction on the structure of $s$.

\HB

\section{An Intermezzo: Constraint Propagation for Linear Constraints}
\label{sec:intermezzo}

Even though we focus here on arithmetic constraints on integer intervals, it is helpful to
realize that the integer interval arithmetic is also useful to define in a succinct way
the well-known rules for constraint propagation for linear constraints
(studied in detail in \cite{HS03}).
To this end consider first a constraint $\Sigma_{i=1}^{n} a_i \cdot x_i = b$, where
$n \geq 0$, $a_1, \LL ,a_n$ are non-zero integers, $x_1, \LL ,x_n$ are
different variables, and $b$ is an integer.
We rewrite this constraint $n$ times, each time isolating one variable, to obtain an extended arithmetic
expression for each variable $x_j$. To each of these extended arithmetic
expressions we apply then the $\opint$ operation of
Subsection~\ref{subsec:correctness}, which yields an update for the domain
of the corresponding variable $x_j$.
To reason about this procedure we can use the following rule parametrized by $j \in [1..n]$:

\begin{center}
\emph{LINEAR EQUALITY}
\[
\frac
{\p{\Sigma_{i=1}^{n} a_i \cdot x_i = b}{x_1 \in D_1, \LL, x_j\in D_j, \LL, x_n \in D_n}}
{\p{\Sigma_{i=1}^{n} a_i \cdot x_i = b}{x_1 \in D_1, \LL, x_j\in D'_j, \LL, x_n \in D_n}}
\]
\end{center}
where 
\(
D'_j := D_j \cap \opint\Big{(}(b - \Sigma_{i \in [1..n] - \{j\}} a_i \cdot x_i)/a_j\Big{)}
\).

Note that by virtue of Note \ref{note:1}
\[
D'_j = D_j \cap (b - \Sigma_{i \in [1..n] - \{j\}} \opint(a_i \cdot D_i))/a_j.
\]

To see that this rule preserves equivalence,
first note that all our reduction rules compute the domain updates via
intersection with the original domain, preventing that domains are extended
by their application.
Further, suppose that for some $d_1 \in D_1, \LL, d_n \in D_n$ we have
$\Sigma_{i=1}^{n} a_i \cdot d_i = b$. Then 
for $j \in [1..n]$ we have
\[
d_j = (b - \Sigma_{i \in [1..n] - \{j\}} a_i \cdot d_i)/a_j
\]
which by the Correctness Lemma \ref{lem:correctness} implies that
\[
d_j \in \opint\Big{(}(b - \Sigma_{i \in [1..n] - \{j\}} a_i \cdot x_i)/a_j\Big{)},
\]
i.e., $d_j \in D'_j$.

Next, consider a constraint $\Sigma_{i=1}^{n} a_i \cdot x_i \leq b$, where
$a_1, \LL ,a_n, x_1, \LL ,x_n$ and $b$ are as above.
To reason about it we can use the following rule parametrized by $j \in [1..n]$:

\begin{center}
\emph{LINEAR INEQUALITY}
\[
\frac
{\p{\Sigma_{i=1}^{n} a_i \cdot x_i \leq b}{x_1 \in D_1, \LL, x_j\in D_j, \LL, x_n \in D_n}}
{\p{\Sigma_{i=1}^{n} a_i \cdot x_i \leq b}{x_1 \in D_1, \LL, x_j\in D'_j, \LL, x_n \in D_n}}
\]
\end{center}
where 
\(
D'_j := D_j \cap (^{\leq} \opint(b - \Sigma_{i \in [1..n] - \{j\}} a_i \cdot x_i)/a_j)
\).

To see that this rule preserves equivalence, suppose that for some
$d_1 \in D_1, \LL, d_n \in D_n$ we have
$\Sigma_{i=1}^{n} a_i \cdot d_i \leq b$. Then
$
a_j \cdot d_j \leq b - \Sigma_{i \in [1..n] - \{j\}} a_i \cdot d_i
$.
By the Correctness Lemma \ref{lem:correctness}
\[
b - \Sigma_{i \in [1..n] - \{j\}} a_i \cdot d_i \in \opint(b - \Sigma_{i \in [1..n] - \{j\}} a_i \cdot x_i),
\]
so by definition
\[
a_j \cdot d_j \in ^{\leq}\opint(b - \Sigma_{i \in [1..n] - \{j\}} a_i \cdot x_i)
\]
and consequently
\[
d_j \in ^{\leq} \opint(b - \Sigma_{i \in [1..n] - \{j\}} a_i \cdot x_i)/a_j.
\]
This implies that $d_j \in D'_j$.

As an alternative to evaluating
\(
\opint(\Sigma_{i \in [1..n] - \{j\}} a_i \cdot x_i)
\)
on every application of the \emph{LINEAR EQUALITY} and \emph{LINEAR INEQUALITY}
rules, we could maintain the interval
\(
\opint(\Sigma_1^n a_i \cdot x_i)
\)
in an auxiliary variable, and subtract
\(
\opint(a_j\cdot x_j)
\)
from it. This corresponds to the two-step propagation described in
\cite{HS03}. If changes to $D_j$ are propagated back
to the auxiliary variable, this does not affect the reduction achieved
by the subsequent applications of the rules, while the number of interval
arithmetic operations involved in the application of a rule becomes constant,
instead of linear in the number $n$ of variables.

\section{Constraint Propagation: Direct Approach}
\label{sec:first}
As a first approach to constraint propagation for arithmetic constraints
on integer intervals, we propose to use the constraints directly, in their
original form. This is an extension of the approach of
Section~\ref{sec:intermezzo} from linear constraints to general arithmetic
constraints, and entails that these constraints are rewritten to isolate
all occurrences of each variable. The resulting extended arithmetic
expressions are then evaluated to obtain updates for the isolated
variables.

The following example illustrates this approach. Consider the constraint 
\[
x^3\cdot y - x \leq 40
\]
and the ranges $x \in [1..100]$ and $y \in [1..100]$.  We can rewrite
it as 
\begin{equation}
  \label{eq:x3}
x \leq \adjfloor{\sqrt[3]{(40 + x) / y}}
\end{equation}
since $x$ assumes integer values.  The maximum value the expression on
the right-hand side can take is $\adjfloor{\sqrt[3]{140}}$, so we
  conclude $x \leq 5$.
By reusing (\ref{eq:x3}), now with the information that $x \in [1..5]$,
we conclude that the maximum value the expression on
the right-hand side of (\ref{eq:x3})
can take is actually $\adjfloor{\sqrt[3]{45}}$,
from which it follows that $x \leq 3$.

In the case of $y$ we can isolate it by rewriting the original
constraint as 
$y \leq 40 / x^3 + 1 /x^2$ 
from which it
follows that $y \leq 41$, since by assumption $x \geq 1$.  So we could
reduce the domain of $x$ to $[1..3]$ and the domain of $y$ to
$[1..41]$.  This interval reduction is optimal, since $x = 1, y = 41$ and $x =
3, y = 1$ are both solutions to the original constraint $x^3\cdot y - x \leq 40$.
So rewriting the constraint as $x\geq x^3\cdot y-40$
does not yield a new lower bound for $x$.

More formally, consider a polynomial constraint
$
\Sigma^{m}_{i=1} m_i = b
$
where $m > 0$, no monomial $m_i$ is an integer, the
power products of the monomials are pairwise
different, and $b$ is an integer.
Suppose that $x_1, \LL, x_n$ are its variables ordered w.r.t.~%
$\prec$.

Select a non-integer monomial $m_{l}$ and assume it is of the form
\[
a \cdot y_{1}^{n_1} \cdot \LL \cdot y_{k}^{n_k},
\]
where $k > 0$, $y_1, \LL, y_k$ are different variables ordered w.r.t.~%
$\prec$,  $a$ is a non-zero integer and $n_1, \LL, n_k$ are
positive integers. So each $y_i$ variable equals to some variable in $\C{x_1, \LL, x_n}$.
Suppose that $y_p$ equals to $x_j$. We introduce the following proof rule:

\begin{center}
\emph{POLYNOMIAL EQUALITY}
\[
\frac
{\p{\Sigma^{n}_{i=1} m_i = b}{x_1 \in D_1, \LL, x_j\in D_j, \LL,  x_n \in D_n}}
{\p{\Sigma^{n}_{i=1} m_i = b}{x_1 \in D_1, \LL, x_j\in D'_j, \LL, x_n \in D_n}}
\]
\end{center}
where
\(
D'_{j} := \opint\left( D_{j} \cap \sqrt[\uproot{4}n_{p}]{\opint\left(({b- \Sigma_{i \in [1..m] - \{{l}\}} m_i})/s\right)}\ \right)
\),
with
\(
s := a \cdot y_{1}^{n_1} \cdot \LL \cdot y_{p-1}^{n_{p-1}} \cdot y_{p+1}^{n_{p+1}} \LL  \cdot y_{k}^{n_k}
\).

To see that this rule preserves equivalence,
choose some $d_1 \in D_1, \LL, d_n \in D_n$.
To simplify the notation, given an extended arithmetic expression 
$t$ denote by  $t'$ the result of evaluating $t$ after 
each occurrence of a variable $x_i$ is replaced by $d_i$.

Suppose that 
$\Sigma_{i=1}^{m} m'_i = b$. Then
\[
d_j^{n_p} \cdot s' = b- \Sigma_{i \in [1..m] - \{{l}\}} m'_i,
\]
so by the Correctness Lemma \ref{lem:correctness} 
applied to $b- \Sigma_{i \in [1..m] - \{{l}\}} m'_i$ and to $s$
\[
d_j^{n_p} \in \opint(b- \Sigma_{i \in [1..m] - \{{l}\}} m_i)/\opint(s).
\]
Hence
\[
d_j \in \sqrt[\uproot{4}n_{p}]{\opint(b- \Sigma_{i \in [1..m] - \{{l}\}} m_i)/\opint(s)}
\]
and consequently
\[
d_j \in \opint\left( D_{j} \cap \sqrt[\uproot{4}n_{p}]{\opint\left(({b- \Sigma_{i \in [1..m] - \{{l}\}} m_i})/s\right)}\ \right)
\]
i.e., $d_j \in D'_j$.
\bigskip

Note that we do not apply $\opint(\cdot)$ to the outcome of the root extraction
operation. For even $n_p$ this means that the second operand of the
intersection can be a union of two intervals, instead of a
single interval. To see why this is desirable, consider the constraint
$x^2-y=0$ in the presence of ranges $x \in [0..10]$, $y \in [25..100]$.
Using the $\opint(\cdot)$ closure of the root extraction we would not be able
to update the lower bound of $x$ to 5.

Next, consider a polynomial constraint $ \Sigma^{m}_{i=1} m_i \leq b.
$ Below we adopt the assumptions and notation used when defining the
\emph{POLYNOMIAL EQUALITY} rule.
To formulate the appropriate rule we stipulate that for the extended
arithmetic expressions $s$ and $t$
\[
\opint((^{\leq}s) / t)\ := \ ^{\geq}Q \cap\ ^{\leq}Q,
\]
with $Q = (^{\leq}\opint(s)) / \opint(t)$.

To reason about this constraint we use the following rule:

\begin{center}
\emph{POLYNOMIAL INEQUALITY}
\[
\frac
{\p{\Sigma^{n}_{i=1} m_i \leq b}{x_1 \in D_1, \LL, x_j\in D_j, \LL, x_n \in D_n}}
{\p{\Sigma^{n}_{i=1} m_i \leq b}{x_1 \in D_1, \LL, x_j\in D'_j,\LL, x_n \in D_n}}
\]
\end{center}
where
\(
D'_{j} := \opint \left(D_{j} \cap \sqrt[\uproot{4}n_{p}]{\opint\left(^{\leq}(b- \Sigma_{i \in [1..m] - \{{l}\}} m_i)/s\right)}\ \right)
\),
with
\(
s := a \cdot y_{1}^{n_1} \cdot \LL \cdot y_{p-1}^{n_{p-1}} \cdot y_{p+1}^{n_{p+1}} \LL  \cdot y_{k}^{n_k}
\).

To prove that this rule preserves equivalence,
choose some $d_1 \in D_1, \LL, d_n \in D_n$.
As above given an extended arithmetic expression 
$t$ we denote by  $t'$ the result of evaluating $t$ when 
each occurrence of a variable $x_i$ in $t$ is replaced by $d_i$.

Suppose that $\Sigma_{i=1}^{m} m'_i \leq b$. Then
\[
d_j^{n_p} \cdot s' \leq b- \Sigma_{i \in [1..m] - \{{l}\}} m'_i.
\]
By the Correctness Lemma \ref{lem:correctness} 
\[
b- \Sigma_{i \in [1..m] - \{{l}\}} m'_i \in \opint(b- \Sigma_{i \in [1..m] - \{{l}\}} m_i),
\] 
so by definition
\[
d_j^{n_p} \cdot s'  \in ^{\leq}\opint(b- \Sigma_{i \in [1..m] - \{{l}\}} m_i).
\]
Hence by the definition of the division operation on the sets of integers
\[
d_j^{n_p} \in ^{\leq}\opint(b- \Sigma_{i \in [1..m] - \{{l}\}} m_i)/\opint(s)
\]
Consequently
\[
d_j \in \sqrt[\uproot{4}n_{p}]{^{\leq}\opint(b- \Sigma_{i \in [1..m] - \{{l}\}} m_i)/\opint(s)}.
\]
This implies that $d_j \in D'_j$.

Note that the set $^{\leq}\opint(b- \Sigma_{i \in [1..m] - \{{l}\}} m_i)$,
which occurs when the expression for $D'_j$ is expanded according to the
above definition of $\opint((^{\leq}s) / t)$, is not an interval. 
So to properly implement this rule we need to extend the implementation of the
division operation discussed in Subsection \ref{subsec:implementation} to the case when
the numerator is an extended interval.

If the sum of the intervals associated with each of the monomials in a
polynomial constraint is maintained in an auxiliary variable, as we discussed
at the end of Section~\ref{sec:intermezzo} for linear constraints,\
then the rules can be applied using a constant number of interval additions.
However, interval
division is not the inverse operation of interval multiplication,
so the same technique cannot be applied to monomials, and 
the number of multiplications and exponentiations will be linear
in the size of the monomial $s$.

In an optimized version of the direct approach we simplify the fractions of
two polynomials by splitting the division over addition and
subtraction and by dividing out common powers of variables and
greatest common divisors of the constant factors.  Subsequently,
fractions whose denominators have identical power products are added.
We used this optimization in the initial example by simplifying
$(40 + x) / x^3$ to $40 / x^3  + 1 / x^2$.  
The reader
may check that without this simplification step we can only deduce that $y
\leq 43$.

To provide details of this optimization, given two monomials $s$ and $t$,
we denote by
\[
[s / t ]
\]
the result of performing this simplification operation on $s$ and $t$.
For example, $[(2 \cdot x^3 \cdot y) / (4 \cdot x^2)]$ equals $(x \cdot y) / 2$,
whereas $[(4 \cdot x^3 \cdot y) / (2 \cdot y^2)]$ equals $(2 \cdot x^3) / y$.

Because the validity of the simplification depends on the sign of the
denominator, we assume that the domains of the variables $y_1,
\LL, y_{p-1}$, $y_{p+1}, \LL, y_n$ of $m_{l}$ do not contain 0. 
For a monomial $s$ involving variables ranging over the integer intervals that do not
contain 0, the set $\opint(s)$ either contains only positive numbers or only negative numbers. 
In the first case we write $\opsign(s) = +$ and in the second case we write
$\opsign(s) = -$.

The new domain of the variable $x_j$ in the \emph{POLYNOMIAL INEQUALITY} 
rule is defined using two sequences $m_0'...m_n'$ and $s_0'...s_n'$ of 
extended arithmetic expressions such that
\[
    m_0' / s_0' = [ b / s ]\ \textrm{and}\
    m_i' / s_i' = - [ m_i / s ]\ \textrm{for $i \in [1..m]$.}
\]
Let $S := \{s_i' \mid i \in [0..m] - \{{l}\} \}$ and 
for an extended arithmetic expression $t \in S$ let 
$I_{t} := \{i \in [0..m] - \{{l}\} \mid s_i' = t\}$. We denote then by 
$p_{t}$ the polynomial $\sum_{i 
\in I_{t}}{m_i'}$. The new domains are then defined by
\[
    D'_{j} := \opint \left( D_{j} \cap \sqrt[\uproot{4}n_{p}]
    {^{\leq}\opint\left( \Sigma_{t \in S}\ p_{t} \oslash t \right)}\ \right)
\]
if $\opsign(s) = +$, and by
\[
    D'_{j} := \opint \left( D_{j} \cap \sqrt[\uproot{4}n_{p}]
    {^{\geq}\opint\left(\Sigma_{t \in S}\ p_{t} \oslash t \right)}\ \right)
\]
if $\opsign(s) = -$.
Here the $\opint(s)$ notation used in the Correctness Lemma
\ref{lem:correctness} is extended to expressions involving the division operator
$\oslash$ on real intervals in the obvious way. We define the $\opint(\cdot)$ operator applied
to a bounded set of reals, as produced by the division and addition
operators in the above two expressions for $D'_j$, to denote the smallest
interval of reals containing that set.

Returning again to the discussion of the two-step propagation technique
of \cite{HS03}, which
we started at the end of Section~\ref{sec:intermezzo}, note that
in this case, the $\opint(\cdot)$ operation is applied after removing
the common power products. For this reason, there is no straightforward
way to calculate
\(
\opint(\Sigma_{t \in S}\ p_{t} \oslash t)
\)
from the sum of all intervals associated with the monomials of a polynomial
constraint.
\section{Constraint Propagation: Partial Decomposition}
\label{sec:second}
As a second approach to constraint propagation for arithmetic constraints
on integer intervals,
we limit our attention to a special type of
polynomial constraints, namely the ones of the form $s \ op \ b$,
where $s$ is a polynomial in which each variable occurs \emph{at most
  once} and where $b$ is an integer. We call such a constraint a
\oldbfe{simple polynomial constraint}.  By introducing auxiliary
variables that are equated with appropriate monomials we can
decompose
any polynomial constraint into a sequence of simple polynomial
constraints.  This allows us also to compute the integer interval
domains of the auxiliary variable from the integer interval domains of
the original variables. We apply then to the simple polynomial
constraints the rules introduced in the previous section.

To see that the restriction to simple polynomial constraints can make
a difference consider the constraint
\[
100x \cdot y - 10y \cdot z = 212,
\]
and ranges $x,y,z\in[1..9]$.
We rewrite it into the sequence
\[
u = x \cdot y, \ v = y \cdot z, \ 100u - 10v = 212,
\]
where $u,v$ are auxiliary variables, each with the domain
$[1..81]$.

It is easy to check that the {\em POLYNOMIAL EQUALITY} rule introduced
in the previous section does not yield any domain reduction when
applied to the original constraint $100x \cdot y - 10y \cdot z = 212$.
In the presence of the discussed optimization the domain of $x$ gets
reduced to $[1..3]$.  

However, if we repeatedly apply the {\em POLYNOMIAL EQUALITY} rule to
the simple polynomial constraint $100u - 10v = 212$, we eventually
reduce the domain of $u$ to the empty set (since this constraint has no
integer solution in the ranges $u,v \in [1..81]$) and consequently can
conclude that the original constraint $100x \cdot y - 10y \cdot z =
212$ has no solution in the ranges $x,y,z\in[1..9]$, without performing
any search.
Note that this effect still occurs if we replace one occurrence of $y$
by a fresh variable with the same domain.

As noted in \cite{clpfd}, decomposing constraints
also prevents the evaluation of subexpressions
whose domains did not change, which may reduce the number of interval
arithmetic operations performed during constraint propagation.
In our case duplicate occurrences
of variables are removed, so the reduction rules additionally become idempotent.
However, this can be seen as a side-effect: rules still update variables that they
depend on, only now this update is indirect, through other variables.

Consider for example the constraint $x^3\cdot y - x \leq 40$ of
Section~\ref{sec:first}. If we rewrite this constraint as $u-x\leq 40$,
with $u=x^3\cdot y$ and $x,y\in[1..100]$, then via $u\leq x+40$ we can set the upper
bound for $u$ to 140. Via $x=\sqrt[3]{u/y}$ we can then set the upper for $x$
to 5. This allows us to set the upper bound for $u$ to 45 via $u\leq x+40$, etc.
From this point of view the auxiliary variables, and the idempotence that
they entail, can be seen as an optimization that prevents the evaluation
of expressions that will not lead to further domain updates.

\section{Constraint Propagation: Full Decomposition}
\label{sec:third}
In this third approach we focus on a small set of `atomic' arithmetic constraints.
We call an arithmetic constraint \oldbfe{atomic} if it is in one of the following two forms:
\begin{itemize}
\item a linear constraint,

\item $x \cdot y = z$.

\end{itemize}

Using appropriate transformation rules
involving auxiliary variables we can
decompose
any arithmetic
constraint to a sequence of atomic arithmetic constraints,
similar to the decomposition  of linear constraints into constraints on
groups of three variables in \texttt{clp(FD)} \cite{clpfd}.
In this transformation, as
with partial decomposition,
the auxiliary variables are
equated with monomials, so we can easily compute their domains.

We explained already in Section \ref{sec:intermezzo} how to
reason about linear constraints.
For a treatment of disequalities see, e.g., \cite{HS03,SS05}.
Next, we focus on the reasoning for the multiplication constraint
$
x \cdot y = z
$
in the presence of the non-empty ranges $x \in D_x$, $y \in D_y$ and $z
\in D_z$.  
To this end we introduce the following three domain reduction rules:

\begin{center}
\emph{MULTIPLICATION 1}
\[
\frac
{\p{x \cdot y = z}{x \in D_x, y \in D_y, z \in D_z}}
{\p{x \cdot y = z}{x \in D_x, y \in D_y, z \in D_z'}}
\]
\medskip
\\
\emph{MULTIPLICATION 2}
\[
\frac
{\p{x \cdot y = z}{x \in D_x, y \in D_y, z \in D_z}}
{\p{x \cdot y = z}{x \in D_x', y \in D_y, z \in D_z}}
\]
\medskip
\\
\emph{MULTIPLICATION 3}
\[
\frac
{\p{x \cdot y = z}{x \in D_x, y \in D_y, z \in D_z}}
{\p{x \cdot y = z}{x \in D_x, y \in D_y', z \in D_z}}
\]
\end{center}
where $D_z':=D_z \cap \opint(D_x \cdot D_y)$,
$D_x':=D_x \cap \opint(D_z/D_y)$, and
$D_y':=D_y \cap \opint(D_z/D_x)$.

The way we defined the multiplication and the division of the integer
intervals ensures that the \emph{MULTIPLICATION} rules \emph{1,2,} and
\emph{3} are equivalence preserving.  Consider for example the
\emph{MULTIPLICATION 2} rule.  Take some $a \in D_x, b \in D_y$ and $c
\in D_z$ such that $a \cdot b = c$. Then $a \in \C{x \in {\cal Z} \mid
  \te z \in D_z \te y \in D_y \: x \cdot y = z}$, so $a \in D_z/D_y$
and a fortiori $a \in \opint(D_z/D_y)$.  Consequently $a \in D_x \cap
\opint(D_z/D_y)$.
Because we also have $D_x \cap \opint(D_z/D_y) \subseteq D_x$,
this shows that the \emph{MULTIPLICATION 2} rule is
equivalence preserving.

The following example shows an interaction between all three
\emph{MULTIPLICATION} rules.

\begin{example} \label{exa:mult} 
Consider the CSP
\begin{equation}
  \label{eq:example}
{\p{x \cdot y = z}{x \in [1..20], y \in [9..11], z \in [155..161]}}.  
\end{equation}

To facilitate the reading we underline the modified domains.
An application of the \emph{MULTIPLICATION 2} rule yields
\[
{\p{x \cdot y = z}{x \in \underline{[16..16]}, y \in [9..11], z \in [155..161]}}
\]
since, as already noted in Subsection \ref{subsec:implementation},
$[155..161]/[9..11]) = [16..16]$, and $[1..20] \cap \opint([16..16]) =
[16..16]$.  Applying now the \emph{MULTIPLICATION 3} rule we obtain
\[
\p{x \cdot y = z}{x \in [16..16], y \in \underline{[10..10]}, z \in [155..161]}
\]
since $[155..161]/[16..16] = [10..10]$ and $[9..11] \cap \opint([10..10]) = [10..10]$.
Next, by the application of the \emph{MULTIPLICATION 1} rule we obtain
\[
\p{x \cdot y = z}{x \in [16..16], y \in [10..10], z \in \underline{[160..160]}}
\]
since $[16..16] \cdot [10..10] = [160..160]$ and $[155..161] \cap \opint([160..160]) = [160..160]$.

So using all three multiplication rules we could solve the CSP (\ref{eq:example}). 
\HB
\end{example}

Now let us clarify why we did not define the 
division of the sets of integers $Z$ and $Y$ by
\[
Z/Y := \C{z/y \in {\cal Z} \mid y \in Y, z \in Z, y \neq 0}.
\]
The reason is that in that case for any set of integers $Z$ we would have
$Z/\C{0} = \ES$. Consequently, if we adopted this definition of the
division of the integer intervals, the resulting \emph{MULTIPLICATION}
\emph{2} and \emph{3} rules would not be anymore equivalence
preserving. Indeed, consider the CSP
\[
\p{x \cdot y = z}{x \in [-2..1], y  \in [0..0],z \in [-8..10]}.
\]
Then we would have $[-8..10]/[0..0]  = \ES$ and consequently
by the \emph{MULTIPLICATION} \emph{2} rule we could conclude
\[
\p{x \cdot y = z}{x \in \ES, y  \in [0..0],z \in [-8..10]}.
\]
So we reached an inconsistent CSP while the original CSP is consistent.

The transformation to atomic constraints can strengthen the reduction.
Consider for example the simple constraint
\[
   w\cdot x\cdot y\cdot z = 24
\]
with $w=4$ and $x,y,z\in[1..4]$. Application of the \emph{POLYNOMIAL EQUALITY}
rule does not reduce any of the domains, but if we replace the constraint
with
\[
u\cdot v=t,\ w\cdot x=u,\ y\cdot z=v
\]
with additional ranges $t=24$, $u\in[4..16]$, and $v\in[1..16]$, then
by application of the \emph{MULTIPLICATION 3} rule to $u\cdot v=t$ we reduce the
domain of $v$ to $[2..6]$. Next, by application of the \emph{MULTIPLICATION 2} rule to
the same constraint we reduce the domain of $u$ to $[4..12]$, and
finally by application of the \emph{MULTIPLICATION 3} rule to $w\cdot x=u$ we
reduce the domain of $x$ from $[1..4]$ to $[1..3]$. Note, however, that this
effect depends on the decomposition. If we had decomposed the constraint as
\[
  z \cdot ( y \cdot ( x\cdot w ) ) = 24
\]
with an auxiliary variable introduced for each pair of matching brackets,
then we would not have
been able to reduce any of the domains of $x$, $y$, and $z$.

In the remainder of the paper we will also consider variants of the full
decomposition approach where we allow squaring and exponentiation as atomic
constraints. For this purpose we explain the reasoning for the constraint
$
x = y^n
$
in the presence of the non-empty ranges $x \in D_x$ and $y \in D_y$, and for $n>1$.
To this end we introduce the following two rules:
\begin{center}
\emph{EXPONENTIATION}
\[
\frac
{\p{x = y^n}{x \in D_x, y \in D_y}}
{\p{x = y^n}{x \in D_x', y \in D_y}}
\]
\\
\medskip
\emph{ROOT EXTRACTION}
\[
\frac
{\p{x = y^n}{x \in D_x, y \in D_y}}
{\p{x = y^n}{x \in D_x, y \in D_y'}}
\]
\end{center}
where $D_x':=D_x \cap \opint(D^n_y)$, and
$D_y':=\opint( D_y \cap \sqrt[n]{D_x})$.

To prove that these rules are equivalence preserving suppose that for
some \mbox{$a\in D_x$} and $b \in D_y$ we have $a = b^n$. Then $a \in
D_y^n$, so $a \in \opint(D_y^n)$ and consequently $a \in D_x \cap
\opint(D_y^n)$.
Also $b \in \sqrt[n]{D_x}$, so
$b \in D_y \cap \sqrt[n]{D_x}$, and consequently
$b \in \opint( D_y \cap \sqrt[n]{D_x})$.

With exponentiation as an atomic constraint, full decomposition leads
to idempotent rules, and the discussion at the end of Section~\ref{sec:second}
applies.

\section{Relation to Hull and Box Consistency}
\label{sec:box}

In this section we relate the three approaches introduced above to the
well-known methods for constraint propagation of arithmetic
constraints on real variables, whose domains are represented by
floating-point intervals. An overview of these methods is
provided in \cite{collaviza99comparing}. \oldbfe{Floating-point intervals} are
intervals of reals, with bounds from a finite set
$\mathcal{F}\subseteq \mathcal{R}\cup\{-\infty,\infty\}$ of
floating-point numbers that contains representations $-\infty$ and
$\infty$ for plus and minus infinity.  For floating-point intervals,
the counterpart of the $\opint(\cdot)$ operation is the \oldbfe{hull}
of a set of real numbers defined as the smallest floating-point
interval containing the set.  Ideally, for an arithmetic constraint
$c$ on the variables $x_1, \LL, x_n$ with respective floating-point
interval domains $D_1, \LL, D_n$ we would like to enforce \oldbfe{hull
  consistency},
which entails that for all $i \in [1..n]$
\[
\begin{array}{l@{\ }l}
D_i = \ophull ( x_i\in\mathcal{R} \mid &
   \exists x_1\in D_1,\ldots, x_{i-1}\in D_{i-1},
           x_{i+1} \in D_{i+1}, \ldots, x_n\in D_n \\
         & (x_1,\ldots,x_n) \in C ).
\end{array}
\]

However, no efficient procedure exists for enforcing hull consistency on
arbitrary arithmetic constraints. Therefore, the natural approach is to first
decompose constraints into atomic constraints, each containing a single
arithmetic operation. Maintaining hull consistency for the decomposed
constraints can be done efficiently, using proof rules similar to the
ones that we introduced, but hull consistency for the resulting decomposed CSP is a
weaker notion of consistency than hull consistency for the original CSP.

Our full decomposition approach can be seen as the integer interval equivalent of
the method for computing hull consistency for a decomposition that we
just described, with the exception that linear constraints are not
decomposed further. In the floating-point case, because of the
accumulation of the rounding errors, the characterization of the
resulting form of constraint propagation in terms of hull consistency
is possible only if all constraints, including linear constraints, are
decomposed into single-operator constraints.

To illustrate this consider the constraint $x+y+z=w$ with the variables ranging
over the floating-point intervals $D_x, D_y, D_z$ and $D_w$.
When we evaluate $D_x+D_y+D_z$ using the floating-point interval arithmetic
to compute an update for $D_w$,
we have three options which two intervals to add first.
Because the floating-point addition is non-associative, we actually compute
the hull of a decomposition that has an extra variable added for
either $x+y$, $x+z$ or $y+z$, and the resulting interval is potentially a proper superset
of $\ophull(D_x+D_y+D_z)$. Moreover, different rewritings of the
constraint correspond to different decompositions, and although this need not
be a problem in practice, the resulting form of local consistency is no longer clearly
defined.

In contrast, for integer intervals, we do not need to deal with the accumulation of the rounding
errors and the linear constraints can be left intact.
Our other two approaches can be seen as variants of the full decomposition
approach that exploit this property further:
for partial decomposition we
allow more than one multiplication per proof rule, and in the direct
approach the decomposition is not made explicit at all.
Apart from these variations, all three approaches are the same in one important
aspect: multiple occurrences of the same variable are treated as different variables.
To illustrate this, consider the constraint $x^3+x=0$, with $x\in[-1..1]$.
While $x=0$ is the unique solution, none of our three approaches will be able
to reduce the domain of $x$. The reason is that the two occurrences of
$x$ are essentially treated as different variables in the reduction rules.
This problem is known as the \oldbfe{dependency problem} of interval arithmetic.

In the context of constraints on reals \cite{benhamou94revisited}
proposed to deal with the dependency problem using the notion of box
consistency.  It is a weaker notion of local consistency than hull
consistency, but is potentially stronger than hull consistency for the
decomposition of a constraint into atomic constraints (see, e.g.,
\cite{collaviza99comparing}). Enforcing box consistency, as described
in \cite{numerica}, consists of fixing the domains of all variables
except one, and then narrowing the domain of this variable by
iteratively instantiating it with subdomains at the boundary of the
original domain, each time verifying consistency of the constraint in
the presence of the domains of the other variables, and subtracting
the subdomain from the original domain if the instantiation leads to a
failure.

The second step of the `trial-and-prune' procedure for enforcing box
consistency that we just sketched can be implemented by enforcing hull
consistency on a decomposition of the original constraint. So the
procedure for enforcing box consistency can be seen as consisting of a
number of procedures including the one that enforces hull consistency.
One could apply the same technique to the arithmetic constraints on
integer intervals, replacing the enforcement of hull consistency by
one of our approaches to constraint propagation.  This would lead to
an integer equivalent of the box consistency.  The efficiency of the
resulting procedure depends on the choice of the underlying approach
to constraint propagation, which provides another argument for the
efficiency analysis of the approaches here considered.

\section{A Characterization of the \emph{MULTIPLICATION} Rules}
\label{sec:characterization}

It is useful to reflect on the effect of the proof rules used to
achieve constraint propagation.  In this section, by way of example,
we focus on the \emph{MULTIPLICATION} rules and characterize their
effect using the notion of bounds consistency as defined in \cite{MS98},
limited to integer intervals.
Let us recall first the definition that we adopt here to the
multiplication constraint.  Given an integer interval $[l..h]$ we
denote by $[l,h]$ the corresponding real interval.

\begin{definition} \label{def:bounds} 
The CSP $\p{x \cdot y = z}{x \in [l_x.. h_x], y \in [l_y.. h_y],z \in [l_z.. h_z]}$
is called \oldbfe{bounds consistent} if

\begin{itemize}
\item $\fa a \in \C{l_x, h_x} \ \te b \in [l_y, h_y] \ \te c \in [l_z, h_z] \ a \cdot b = c$,

\item $\fa b \in \C{l_y, h_y} \ \te a \in [l_x, h_x] \ \te c \in [l_z, h_z] \ a \cdot b = c$,

\item $\fa c \in \C{l_z, h_z} \ \te a \in [l_x, h_x] \ \te b \in [l_y, h_y] \ a \cdot b = c$.
\HB
\end{itemize}
\end{definition}

The following result entails that
the \emph{MULTIPLICATION} rules will not reduce a CSP beyond bounds consistency.

\begin{theorem}[Bounds consistency] \label{thm:multi}
  Suppose a CSP $\p{x \cdot y = z}{x \in D_x, y \in D_y,z \in D_z}$
  with the integer interval domains is bounds consistent. Then it is
  closed under the applications of the MULTIPLICATION 1,2 and 3 rules.
\end{theorem}
\Proof
See the Appendix.
\HB
\bigskip

This theorem shows that the \emph{MULTIPLICATION} rules entail a notion of
local consistency, say $M$-consistency, that is implied by bounds
consistency.
However, $M$-consistency does not imply bounds consistency. Here is an example.
Consider the CSP
\[
\p{x \cdot y = z}{x \in [-2..1], y  \in [-3..10],z \in [8..10]}.
\]
It is not bounds consistent, since for $y = -3$ no real values
$a \in [-2,1]$ and $c \in [8,10]$ exist such that 
$a \cdot (-3) = c$. Indeed, it is easy to check that
\[
\C{y \in {\cal R} \mid \te x \in [-2,1] \: \te z \in [8,10] \: x \cdot y = z} = 
(- \infty, -4] \cup [8, \infty).
\]

However, this CSP is closed under the applications of the
\emph{MULTIPLICATION} \emph{1, 2} and \emph{3} rules since
\begin{itemize}
\item $[8..10] \sse \opint([-2..1] \cdot [-3..10])$, as $\opint([-2..1] \cdot [-3..10]) = [-20..10]$,

\item $[-2..1] \sse \opint([8..10]/[-3..10])$ as $\opint([8..10]/[-3..10]) = [-10..10]$, and

\item $[-3..10] \sse \opint([8..10]/[-2..1])$ as $\opint([8..10]/[-2..1]) = [-10..10]$.
\end{itemize}

The following result clarifies that this example identifies the
only cause of discrepancy between $M$-consistency
and bounds consistency.
Here, given an integer interval $D := [l .. h]$
we define $\langle D \rangle := \C{x \in {\cal Z} \mid l < x < h}$.

\begin{theorem}[Bounds consistency 2] \label{thm:multi2}
  Consider a CSP $\phi := \p{x \cdot y = z}{x \in D_x, y \in
    D_y,z \in D_z}$ with non-empty integer interval domains and such that
  \begin{equation}
    \label{eq:condition}
    \mbox{$0 \in \langle D_x \rangle \cap \langle D_y \rangle$ implies $0 \in D_z$}.
  \end{equation}
Suppose $\phi$ is closed under the applications of the MULTIPLICATION 1,2 and 3
  rules.  Then it is bounds consistent.
\end{theorem}
\Proof
See the Appendix.
\HB
\bigskip

Consequently the \emph{MULTIPLICATION} rules only fail to enforce bounds
consistency for the constraint $x\cdot y=z$ in case the domains of
$x$ and $y$ are both of the form $[l..h]$, with $l<0$ and $h>0$ while
$z$ can assume either only positive numbers, or only negative numbers.
Because the zeroes in the domains of $x$ and $y$ do not contribute to
any solution, we can remedy this effect by temporarily splitting these
domains in a positive interval and a negative interval. Bounds
consistency for the constraint $x\cdot y=z$ is then achieved by
applying the \emph{MULTIPLICATION} rules to the resulting subproblems,
and updating the domain of each variable with the $\opint(\cdot)$
closure of the union of its domain in these subproblems. 

In \cite{SS05} similar rules to our \emph{MULTIPLICATION} rules are
defined that apply this technique directly.  They were defined without
the use of interval arithmetic. It is also shown there that the
\emph{LINEAR EQUALITY} and \emph{LINEAR INEQUALITY} rules enforce
bounds consistency.
\section{Implementation Details}
\label{sec-experiments}

\subsection{Weak Division}
\label{subsec: weak}

We already mentioned in Section \ref{sec:interval} that the division operation on 
the intervals does not admit an efficient implementation.
The reason is that the $\opint(\cdot)$ closure of the interval division $[a..b] / [c..d]$ 
requires an auxiliary computation in case when $0 \not \in [c..d]$.
The preprocessing of $[c..d]$ becomes impractical for small intervals $[a..b]$,
and large $[c..d]$, occurring for example for the constraint
$\prod_{i=1}^{n}x_i = \prod_{i=1}^{n}i$, of the benchmark problem mentioned
in Subsection~\ref{subsec:motivation}.
This can be remedied by using the following variant of the division operation.
We call it \oldbfe{weak division} since it yields a larger set (and so is `weaker').

\[
[a..b] : [c..d] :=    
\left\{ 
\begin{tabular}{ll}
$[\adjceiling{\opmin(A)} .. \adjfloor{\opmax(A)}]$        &  \mbox{if $0 \not\in [c..d]$, or}\\
                                                    &  \mbox{\phantom{if} $0 \notin [a..b]$} and $0\in\{c,d\}$ and $c<d$, \\
$[a..b] / [c..d]$                                   &  otherwise
\end{tabular}
\right . 
\]
where $A = \C{a/c', a/d', b/c', b/d'}$, and $[c'..d']=[c..d]-\{0\}$.

Then $\opint([a..b] : [c..d])$ can be computed by a straightforward case analysis already
used for $\opint([a..b] / [c..d])$ but now without any auxiliary computation.
The weak division operator gives rise to the following versions of
the \emph{MULTIPLICATION} rules 2 and 3:
\begin{center}
\emph{MULTIPLICATION 2w}
\[
\frac
{\p{x \cdot y = z}{x \in D_x, y \in D_y, z \in D_z}}
{\p{x \cdot y = z}{x \in D_x', y \in D_y, z \in D_z}}
\]
\medskip
\\
\emph{MULTIPLICATION 3w}
\[
\frac
{\p{x \cdot y = z}{x \in D_x, y \in D_y, z \in D_z}}
{\p{x \cdot y = z}{x \in D_x, y \in D_y', z \in D_z}}
\]
\end{center}
where $D_x':=D_x \cap \opint(D_z : D_y)$,
and $D_y':=D_y \cap \opint(D_z : D_x)$.

In the assumed framework based on constraint propagation and tree search,
all domains become eventually singletons or empty sets.
It can easily be verified that both division operations
are then equal, i.e., $[a..b]:[c..d] = [a..b]/[c..d]$,
for $a\geq b$ and $c\geq d$.
For this reason, we can safely replace any of the reduction rules introduced
in this paper,
notably \emph{POLYNOMIAL EQUALITY}, \emph{POLYNOMIAL INEQUALITY}, and
\emph{MULTIPLICATION} 2 and 3, by their counterparts based on the weak division.
For the \emph{MULTIPLICATION} rules specifically, the following theorem
states that both sets of rules actually achieve the same constraint
propagation.

\begin{theorem}[\emph{MULTIPLICATION}] \label{thm:multi3}
  A CSP $\p{x \cdot y = z}{x \in D_x, y \in D_y,z \in D_z}$
  with the integer interval domains is closed under the applications
  of the MULTIPLICATION 1, 2 and 3 rules iff it is closed under the
  applications of the MULTIPLICATION 1, 2w and 3w rules.
\end{theorem}
\Proof
See the Appendix.
\HB

\bigskip
 
Let us clarify now the relation between the \emph{MULTIPLICATION}
rules and the corresponding rules based on real interval arithmetic
combined with the rounding of the resulting real intervals inwards to
the largest integer intervals.
The CSP $\p{x \cdot y = z}{x \in [-3..3], y \in [-1..1], z \in [1..2]}$,
which we already discussed in the introduction,
shows that these approaches yield different results.
Indeed, using the \emph{MULTIPLICATION} rule 2 we can reduce the
domain of $x$ to $[-2..2]$, while the latter approach yields no
reduction.
 
On the other hand, the applications of the \emph{MULTIPLICATION} rules
2w and 3w to $\p{x \cdot y = z}{x \in D_x, y \in D_y,z \in D_z}$ such
that $\opint(D_z : D_x) \neq \opint(D_z / D_x)$ and $\opint(D_z : D_y) \neq
\opint(D_z / D_y)$ (so in cases when the use of the weak interval
division differs from the use of the interval division) do coincide
with the just discussed approach 
based on real interval arithmetic and inward rounding. This is a
consequence of the way the multiplication and division of real
intervals are defined, see \cite{HJvE01}.
However, as we already stated in the introduction, we believe that
the limited precision of floating-point interval arithmetic, and the
modest overhead of arbitrary length integers justify a separate
implementation of these rules for arithmetic constraints on integer intervals.

\subsection{Implementation}

\subsubsection*{Platform}

Our experiments were performed using OpenSolver~\cite{pz03os},
an experimental constraint solver based on constraint propagation and
tree search. OpenSolver can be configured by software plug-ins in a number
of predefined categories, corresponding to different aspects of
constraint propagation and tree search,
which makes it particularly well-suited for carrying out comparative
studies of implementations of constraint solvers.
The categories of plug-ins that are relevant
for the experiments reported here are:
\begin{itemize}
   \item variable domain types, which implement the domains of variables,
   \item domain reduction functions (DRFs), which correspond to the reduction
       rules,
   \item schedulers of DRFs, which determine the order in which the DRFs are
       applied,
   \item branching strategies, which split the search tree after constraint
      propagation has terminated, and
   \item several categories corresponding to different aspects
      of a search strategy that determine how to traverse a search tree.
\end{itemize}

All experiments were performed using the \texttt{IntegerInterval}
variable domain type plug-in. Domains of this type consist of an
indication of the type of the interval (bounded, unbounded,
left/right-bounded, or empty), and a pair of arbitrary length
integer bounds. This plug-in, and the DRFs operating
on it are built using the already mentioned GNU MP library, which
provides arbitrary length integers and arithmetic operations on them,
including operations for rounding the
outcome of divisions and root extractions in the desired direction.

The branching strategy that we used selects variables using a chronological
ordering in which the auxiliary variables come last.
The domain of the selected variable is split into two
subdomains using bisection, so the resulting search trees are binary
trees.  In all experiments we searched for all solutions, traversing
the entire search tree by means of depth-first leftmost-first
chronological backtracking.

For the experiments in this paper a DRF plug-in has been developed that
implements the domain reduction rules discussed in the previous sections.
Every constraint of a CSP is enforced by a number of instantiations of this
DRF: one for each variable occurrence.

The scheduler plug-in that we used in the experiments maintains a flag
per DRF, indicating whether the DRF is pending application or not.
Initially, all DRFs are pending application. If the application of a DRF
(or the branching strategy)
modifies the domains of one or more variables, all DRFs whose output
depends on these variables become pending application. Since in general
---as illustrated by the example at the beginning of Section~\ref{sec:first}---
the DRFs are non-idempotent, this may include the DRF that has just been
applied. By default, the scheduler plug-in keeps cycling through the
set of DRFs for a given CSP in a specified order, applying those DRFs that
are pending application. The cycling stops when no DRF is pending application,
or when the domain of a variable becomes empty.

\subsubsection*{Scheduling of Reduction Rules}

It was already shown in \cite{cai92wallace} that controlling
the order in which variables are updated can improve the efficiency of
constraint propagation algorithms, and for this purpose, our scheduler
plug-in can be supplied with a \oldbfe{schedule}.
Such a schedule is a sequence of indices into the set of DRFs that
describes the order in which the scheduler will visit them, as an
alternative to cycling.
This is used in combination with full and partial decomposition, where
we distinguish \oldbfe{user constraints} from the constraints that are
introduced to define the values of auxiliary variables.
Before considering for execution a DRF $f$ that is part of the implementation
of a user constraint, we make sure that all auxiliary variables that $f$
relies on are updated. For this purpose, the indices of the
DRFs that update these variables precede the index of $f$ in the schedule.
If $f$ can change the value of an auxiliary variable, its index is followed
by the indices of the DRFs that propagate back these changes to the
variables that define the value of this auxiliary variable.

For example, rewriting $x^3\cdot y - x \leq 40$ to simple constraints
introduces an auxiliary variable $u$, which is equated with $x^3\cdot y$.
This leads to five reduction rules: one for each occurrence of a variable
after the rewriting step. We number these reduction rules
as follows, where we underline in the constraint
the variable that is updated by the rule:

\bigskip
\noindent
\hspace{\stretch{3}} 1. $\underline{u} = x^3\cdot y$
\hspace{\stretch{1}} 2. $u = \underline{x}^3\cdot y$ 
\hspace{\stretch{1}} 3. $u = x^3\cdot \underline{y}$
\hspace{\stretch{3}}

\hspace{\stretch{3}} 4. $\underline{u}-x\leq 40$
\hspace{\stretch{1}} 5. $u-\underline{x}\leq 40$
\hspace{\stretch{3}}

\bigskip
\noindent
The fragment of the generated schedule that corresponds
to enforcing the constraint $x^3\cdot y - x \leq 40$ is then 4,2,3,1,5. Rules 4 and 5
correspond to the original inequality, but rule 4 potentially modifies $u$,
so in the schedule, rule 4 is followed by rules 2 and 3, that propagate any
changes to $u$ back to $x$ and $y$. Before considering rule 5 for application,
the schedule specifies that first rule 1 should be considered, so that
any changes to the domains of $x$ and $y$ are propagated to the domain of $u$.

To see that an appropriate scheduling of the rules can be
beneficial compared to cycling through the rules, suppose that all rules
are pending application, and that $D_x=D_y=[1..100]$, and $D_z=\mathcal{Z}$.
If we iterate the rules in their original order 1,2,3,4,5 then
we first reduce $D_u$ to $[1..100^4]$ by means of rule 1.
Next, rules 2 and 3 are executed without making any changes.
Rule 4 then reduces $D_u$ to $[1..140]$, which makes rules 2 and 3 pending
application again. Next, rule $5$ is executed
without reducing $D_x$. Because $x$ and $y$ have not changed, rule 1 is not
set to pending application, and rule 2 is the first rule that is applied in the
second cycle, which reduces $D_x$ to $[1..5]$. If use the generated schedule
4,2,3,1,5 instead, the same reduction is achieved immediately after applying
the first two rules, instead of the six rules that are applied if we just
cycle through the rules.

For full decomposition, there can be hierarchical dependencies between auxiliary
variables.
Much like the HC4revise procedure of~\cite{granvilliers99revising},
the generated schedule then specifies a bottom-up traversal of this hierarchy
in a forward evaluation phase, and a top-down traversal in a backward
propagation phase. These phases are performed before and after applying
a DRF of a user constraint, respectively.
In the forward evaluation phase, the DRFs that are executed correspond to
the \emph{MULTIPLICATION 1} and \emph{EXPONENTIATION} rules. The DRFs of the
backward propagation phase correspond to the \emph{MULTIPLICATION 2} and \emph{3\/},
and \emph{ROOT EXTRACTION} rules.
The HC4revise procedure is part of the HC4 algorithm, which enforces
hull consistency for constraints on the reals using an implicit decomposition.
For a discussion of this algorithm in the context of controlled constraint
propagation, see \cite{sac05goualard}.

\subsubsection*{Constraint Rewriting}

The proposed approaches were implemented by first rewriting arithmetic constraints
to polynomial constraints, and then to
a sequence of DRFs that correspond to the rules of the approach used.
We implemented the following variants:

\begin{itemize}
\item[\appdu](direct, unoptimized): the direct approach, discussed in Section \ref{sec:first}, where
we isolate all variable occurrences in the original constraints without
decomposing them first;

\item[\appdo](direct, optimized): the optimization of the direct approach, discussed at the end of
Section \ref{sec:first}, which involves dividing out common powers of variables
in the extended arithmetic expressions that arise from isolating the
variable occurrences;

\item[\apppu](partial, unoptimized): partial decomposition into simple constraints, as
  discussed in Section \ref{sec:second}.
  The decomposition is implemented
  by introducing an auxiliary variable for every nonlinear
  power product.
  This procedure may introduce more auxiliary variables than necessary;

\item[\apppo](partial, optimized): an optimized version of variant \apppu, where we stop
  introducing auxiliary variables as soon as the constraints contain
  no more duplicate occurrences of variables;

\item[\appfm](full, multiplication): full decomposition into atomic constraints,
  as discussed in Section \ref{sec:third},
  allowing only linear constraints and multiplication as atomic constraints;
\item[\appfs](full, squaring): idem, but also allowing $x = y^2$ as an atomic constraint;

\item[\appfe](full, exponentiation): idem, allowing $x = y^n$ for all $n >1$ as an atomic constraint.

\end{itemize}
If the distinction between the different variants of an approach is
irrelevant, we will sometimes omit the subscripts to the names
\appd, \appp, and \appf.

Full and partial decomposition
are implemented as a rewrite step, where the auxiliary
variables are introduced.
The resulting CSP is then rewritten using the direct approach.
During the first rewrite step the hierarchical relations between the
auxiliary variables are recorded, and the schedules are generated as a part
of the second rewrite step.
For variants \apppo and \appf
the question of which auxiliary variables to introduce
is an optimization problem in itself. Some choices result in more
auxiliary variables than others. We have not treated this issue as an
optimization problem but relied on the (somewhat arbitrary) heuristics
described below.
For this reason we have to consider the possibility that performance of
variants \apppo and \appf
can be further improved because in our experiments
we used a suboptimal decomposition. The heuristics are as follows.
\begin{itemize}
   \item For
      variant \apppo
      we replace nonlinear power products from left to right,
      so the rightmost nonlinear term of a polynomial constraint is always left intact.
   \item For the
      full decomposition
      approach, nonlinear power products are processed
      in the order in which they occur in the problem statement,
      after normalization to polynomial constraints.
      On the first occurrence of a nonlinear power product, we start
      introducing auxiliary variables for terms that divide the power product
      by multiplying or exponentiating existing variables, and keep doing
      so  until
      we have introduced an auxiliary variable that corresponds to the
      full power product. When there are several choices for which existing
      variables to multiply or exponentiate, we introduce an auxiliary
      variable for a term with the largest possible sum of exponents,
      thereby giving preference to exponentiation over
      multiplication, insofar as it is allowed by the
      variant.
      For variant \appfe we first introduce auxiliary variables for all
      exponentiations in the power product.
      For variant \appfs, we first introduce auxiliary variables for all
      exponentiations that divide the power product, and
      whose exponent is a power of 2. Unused auxiliary variables are deleted
      at a later stage.
\end{itemize}

To illustrate the latter heuristic, suppose we want to introduce an
auxiliary variable for the term $x^5\cdot y^3\cdot z$.
If we allow exponentiation,
we start by introducing auxiliary variables $u_1$ and $u_2$
for the exponentiations in the term, and constrain them as follows:
$u_1=x^5$, $u_2=y^3$. Next we can introduce an auxiliary variable
$u_3$ for $x^5\cdot y^3$, $x^5\cdot z$, or $y^3\cdot z$ by adding a constraint
that multiplies two of the variables $u_1$, $u_2$, and $z$.
Because the sum of exponents is highest for the first option, we add
$u_3=u_1\cdot u_2$. Finally $u_4$ is introduced to replace the original
term: $u_4 = u_3\cdot z$. With only squaring allowed, we would be
making these decisions in the presence of auxiliary variables for
$x^2$, $x^4$, and $y^2$, where $x^4$ is obtained by squaring $x^2$.
In this case, the first auxiliary variable introduced would be for
$x^4\cdot y^2$.
With only multiplication allowed, after introducing $u_1=x\cdot x$ and
$u_2=u_1\cdot u_1$, we would be expanding the term be repeatedly
multiplying it with $x$, $y$, or $z$.

Except for the optimized version of the direct approach,
our current implementation can be optimized further by adopting
the two-step propagation of linear constraints described in
\cite{HS03}, as discussed at the end of
Section~\ref{sec:intermezzo}. Because linear constraints are never
decomposed, the effect is essentially the same for all
alternatives
that we discussed, so we have not considered this technique in our
evaluation.
\section{Experiments}
\label{sec:experiments}
\subsection{Problems}
\label{subsec:problems}
For evaluating the alternative approaches, we used the integer problems
described below. Problems with only integer variables and arithmetic
constraints are rare in practice, and in that sense, our benchmark problems
are artificial, but they serve well to generate a purely integer workload.
The approach that works best on these problems can also be expected to work
well in a hybrid setting, where integer variables and arithmetic constraints
are mixed with other types of variables and constraints. In that case,
only a fraction of the workload will be devoted to integer arithmetic,
but with the results of this study we can be confident that we are dealing
with these constraints in an efficient way.

\paragraph*{Cubes}
The problem is to find all natural numbers
$n \leq 10^5$
that are a sum of four different cubes, for example
\begin{displaymath}
   1^3 + 2^3 + 3^3 + 4^3 = 100.
\end{displaymath}
This problem is modeled as follows:
\begin{displaymath}
   \begin{array}{l}
   \langle
   1 \leq x_1,\ x_1 \leq x_2-1,\ x_2 \leq x_3-1,\ x_3 \leq x_4-1,\ x_4 \leq n,\\
   \phantom{\langle} x_1^3 + x_2^3 + x_3^3 + x_4^3 = n ;\
   n \in [1..10^5],\ x_1,x_2,x_3,x_4 \in {\cal Z}
   \rangle
   \end{array}
\end{displaymath}

\paragraph*{Opt}
We are interested in finding a solution to the constraint
$x^3 + y^2 = z^3$ in the integer interval $[1..10^5]$ for which
the value of $2x\cdot y - z$ is maximal.

\paragraph*{Fractions}
This problem is taken from~\cite{SS02}: find distinct
nonzero digits such that the following equation holds:
\begin{displaymath}
   \frac{A}{BC} + \frac{D}{EF} + \frac{G}{HI} = 1
\end{displaymath}
There is a variable for each letter. The initial domains are $[1..9]$.
To avoid symmetric solutions an ordering is imposed:
\begin{displaymath}
   \frac{A}{BC} \geq \frac{D}{EF} \geq \frac{G}{HI}
\end{displaymath}
Also two redundant constraints are added:
\begin{displaymath}
   3 \frac{A}{BC} \geq 1 \qquad \textrm{and} \qquad
   3 \frac{G}{HI} \leq 1
\end{displaymath}
Because division is not present in our arithmetic expressions, the
above constraints are multiplied by the denominators of the fractions
to obtain arithmetic constraints.
We studied a representation for this problem using
one equality and four inequalities for
the ordering and the redundant constraints, and
36 disequalities $A \neq B$, $A \neq C$, ..., $H \neq I$.

\paragraph*{Kyoto}
The problem (see \cite{cyberteaser})
is to find the number $n$ such that
the alphanumeric equation

\begin{centering}
\begin{tabular}{llllll}
  & K & Y & O & T & O \\
  & K & Y & O & T & O \\
+ & K & Y & O & T & O \\
\hline
  & T & O & K & Y & O
\end{tabular}\\
\end{centering}
\noindent
has a solution in the base-$n$ number system. Our representation uses a variable for
each letter and one variable for the base number.
The variables $K$ and $T$ may not be zero.
There is one large constraint for the addition, 6 disequalities
$K\neq Y$ ... $T \neq O$ and
four constraints stating that the
individual digits $K, Y, O, T$, are smaller than the base number. To spend some
CPU time, we searched base numbers 2..100.

\paragraph*{Sumprod}
This is the problem cited in Subsection~\ref{subsec:motivation}, for $n=14$.
We use the following representation:
\[
\begin{array}{l}
   \langle
           x_1+\ldots+x_n = c_1+\ldots+c_n,\\
   \phantom{\langle}
           x_1\cdot\ldots\cdot x_n = c_1\cdot\ldots\cdot c_n,\\
   \phantom{\langle}
           x_1\leq x_2, x_2\leq x_3, \ldots, x_{n-1}\leq x_n\ ;\ \\
   \phantom{\langle}
           x_1,\ldots,x_n\in[1..n],\\
   \phantom{\langle}
           c_1\in\{1\},c_2\in\{2\},\ldots,c_n\in\{n\} \rangle
\end{array}
\]
For $n=14$, the value of the expression $\prod_{i=1}^{n}i$ equals
$14!$, which exceeds $2^{32}$, and to avoid problems with the input of large
numbers, we used bound variables $c_1,\ldots,c_n$ and constraint propagation
to evaluate it.
\subsection{Results}
\label{subsec:results}
Tables~\ref{tab-elapsed} and~\ref{tab-numbers} compare the
implemented variants of our
approaches on the problems defined in the previous subsection.
The first two columns of table~\ref{tab-elapsed} list
the number of variables and DRFs that were used. Column nodes
lists the size of the search tree, including failures and solutions.
The next two columns list the number of times that a DRF was applied,
and the percentage of these applications that the domain of a variable was
actually modified. For the {\em opt} problem, the DRF that implements the
optimization is not counted, and its application is not taken into account.
The reported CPU times are user time in seconds, as reported by the
UNIX \texttt{time} command on a 1200 MHz Athlon CPU. The last
column compares the performance of our
implementation
to that of \eclipse,
and will be discussed at the end of this section.
{
\setlength\tabcolsep{2pt}
\begin{table}[htbp]
\begin{centering}
\begin{tabular}{|r@{\hspace{1ex}}|@{\hspace{2ex}}r@{\hspace{2ex}}r@{\hspace{2ex}}
                r@{\hspace{2ex}} r@{\hspace{2ex}}
                r@{\hspace{2ex}} r@{\hspace{2ex}}|
                 @{\hspace{1ex}} c@{\hspace{1ex}}|}
\hline
&      &      &       & DRFs\hspace{1ex}      &        &CPU&   \\
& nvar & nDRF & nodes & applied   & \%eff. & (sec.) & \eclipse \\
\hline
\multicolumn{1}{|@{\hspace{1ex}}l|@{\hspace{2ex}}}{\textit{cubes}} & & & & & & & 6.54s\\
\appd,\apppo &    5 &   14 &   169,755 &    1,876,192 &    9.52 &     9.69  & = \\
\apppu,\appfe &    9 &   22 &   169,755 &    2,237,590 &   16.28 &     6.53  & = \\
\appfm &   13 &   34 &   206,405 &    3,011,749 &   20.02 &     8.53  & - \\
\appfs &   13 &   34 &   178,781 &    2,895,717 &   20.62 &     8.80  & - \\
\hline
\multicolumn{1}{|@{\hspace{1ex}}l|@{\hspace{2ex}}}{\textit{opt}} & & & & & & & 5752.70s \\
\appd,\apppo &    4 &    7 &   115,469 &    5,187,002 &   42.16 &    21.55  & + \\
\apppu,\appfe &    8 &   15 &   115,469 &    9,800,017 &   60.00 &    22.75  & + \\
\appfm &   10 &   21 &         ? &            ? &       ? &         ? & - \\
\appfs &   10 &   21 & 5,065,195 &  156,906,444 &   46.49 &   422.93  & - \\
\hline
\multicolumn{1}{|@{\hspace{1ex}}l|@{\hspace{2ex}}}{\textit{fractions}} & & & & & & & 6.90s \\
\appdu &    9 &  154 &    11,289 &    1,193,579 &    3.65 &    15.40  & = \\
\appdo &    9 &  154 &     7,879 &      734,980 &    3.45 &    17.38  & = \\
\apppu &   37 &  210 &    11,289 &    1,410,436 &   23.27 &     4.89  & = \\
\apppo &   32 &  200 &    11,289 &    1,385,933 &   21.65 &     5.25  & = \\
\appf  &   43 &  208 &    11,131 &    1,426,204 &   27.76 &     4.98  & = \\
\hline
\multicolumn{1}{|@{\hspace{1ex}}l|@{\hspace{2ex}}}{\textit{kyoto}} & & & & & & & 302.73s \\
\appdu &    5 &   37 &    87,085 &    3,299,814 &    6.09 &    21.84  & = \\
\appdo &    5 &   37 &    87,085 &    3,288,461 &    5.94 &    44.56  & + \\
\apppu &   13 &   53 &    87,085 &    3,781,514 &   23.02 &    10.93  & = \\
\apppo &   12 &   51 &    87,085 &    3,622,461 &   21.45 &    11.24  & = \\
\appfm &   16 &   60 &    87,087 &    4,276,066 &   26.70 &    10.40  & = \\
\appfs &   16 &   60 &    87,085 &    4,275,957 &   26.70 &    10.39  & = \\
\appfe &   16 &   59 &    87,085 &    3,746,532 &   23.26 &     9.42  & = \\
\hline
\multicolumn{1}{|@{\hspace{1ex}}l|@{\hspace{2ex}}}{\textit{sumprod}} & & & & & & & 23.25s \\
\appd,\apppo &   28 &   82 &   230,233 &   10,910,441 &    7.91 &   102.49  & = \\
\apppu &   30 &   86 &   230,233 &    9,196,772 &    9.39 &    80.59  & = \\
\appf  &   54 &  134 &    55,385 &    3,078,649 &   18.01 &    23.75  & = \\
\hline
\end{tabular}
\\
\end{centering}
\caption{Statistics and comparison with \eclipse}
\label{tab-elapsed}
\end{table}
}

Table~\ref{tab-numbers} lists measured numbers of basic interval operations.
Note that for variant \appdo, there are two versions of the division and addition
operations: one for integer intervals, and one for
intervals of reals of which the bounds are rational numbers
(marked $\mathcal{Q}$).
Columns multI and multF list the numbers of multiplications of two integer
intervals, and of an integer interval and an integer factor, respectively.
These are different operations in our implementation.
{
\setlength\tabcolsep{2pt}
\begin{table}[htbp]
\begin{centering}
\begin{tabular}{|r@{\hspace{1ex}}|@{\hspace{2ex}}r@{\hspace{2ex}}r@{\hspace{2ex}}rlr@{\hspace{2ex}}r@{\hspace{2ex}}rlr@{\hspace{1ex}}|}
\hline
& root & exp & div & & multI & multF & sum & & total \\
\hline
\multicolumn{1}{|@{\hspace{1ex}}l|@{\hspace{2ex}}}{\textit{cubes}} & & & & & & & & & \\
\appd,\apppo  &    1,182 &  4,224 &      0  &  &      0 &   4,756 &  4,245 & & 14,408 \\
\apppu,\appfe  &      180 &    181 &      0  &  &      0 &   4,756 &  4,245 & &  9,363 \\
\appfm  &        0 &      0 &    589  &  &    438 &   4,927 &  4,363 & & 10,317 \\
\appfs  &      192 &    198 &    384  &  &    198 &   4,842 &  4,305 & & 10,121 \\
\hline
\multicolumn{1}{|@{\hspace{1ex}}l|@{\hspace{2ex}}}{\textit{opt}} & & & & & & & & & \\
\appd,\apppo  &    2,299 &  4,599 &  1,443  &  &  1,444 &  11,064 &  5,187 & & 26,037 \\
\apppu,\appfe  &    1,636 &  1,538 &  2,150  &  &    738 &   8,138 &  4,445 & & 18,645 \\
\appfm  &        ? &      ? &      ?  &  &      ? &       ? &      ? & &      ? \\
\appfs  &   21,066 & 18,106 & 54,172  &  & 18,285 & 106,652 & 57,470 & &275,751 \\
\hline
\multicolumn{1}{|@{\hspace{1ex}}l|@{\hspace{2ex}}}{\textit{fractions}} & & & & & & & & & \\
\appdu  &        0 &      0 &    868  &  & 28,916 &  14,238 & 13,444 & & 57,466 \\
\appdo  &        0 &      0 &     51  &  & 11,892 &   8,010 &  6,727 & & 29,584 \\
    &          &        &  1,550  & $\mathcal{Q}$ &        &         &  1,355 & $\mathcal{Q}$&        \\
\apppu  &        0 &      0 &    734  &  &    933 &   4,736 &  4,669 & & 11,071 \\
\apppo  &        0 &      0 &    776  &  &  1,509 &   5,292 &  5,147 & & 12,725 \\
\appf   &        0 &      0 &    693  &  &    339 &   4,835 &  4,769 & & 10,636 \\
\hline
\multicolumn{1}{|@{\hspace{1ex}}l|@{\hspace{2ex}}}{\textit{kyoto}} & & & & & & & & & \\
\appdu  &      735 & 11,041 &  1,963  &  & 13,853 &  10,853 & 13,946 & & 52,390 \\
\appdo  &      735 &  8,146 &    218  &  &  8,955 &  12,516 & 10,592 & & 48,749 \\
    &          &        &  4,310  & $\mathcal{Q}$ &        &         &  3,277 & $\mathcal{Q}$&        \\
\apppu  &      383 &    759 &  1,591  &  &    484 &   5,324 &  7,504 & & 16,044 \\
\apppo  &      383 &    759 &  1,597  &  &  1,360 &   5,756 &  8,008 & & 17,863 \\
\appfm  &        0 &      0 &  1,991  &  &    578 &   5,324 &  7,505 & & 15,398 \\
\appfs  &   $<0.5$ & $<0.5$ &  1,990  &  &    578 &   5,324 &  7,504 & & 15,397 \\
\appfe  &        1 &      1 &  1,554  &  &    484 &   5,324 &  7,504 & & 14,868 \\
\hline
\multicolumn{1}{|@{\hspace{1ex}}l|@{\hspace{2ex}}}{\textit{sumprod}} & & & & & & & & & \\
\appd,\apppo  &        0 &      0 &  4,032  &  &100,791 &  85,419 &149,479 & &339,721 \\
\apppu  &        0 &      0 &  2,186  &  & 27,948 &  81,728 &149,479 & &261,340 \\
\appf   &        0 &      0 &    609  &  &    205 &  25,799 & 46,960 & & 73,573 \\
\hline
\end{tabular}
\\
\end{centering}
\caption{Measured numbers (thousands) of interval operations}
\label{tab-numbers}
\end{table}
}
 
For the \emph{cubes}, \emph{opt}, and \emph{sumprod} problems, the constraints are already
in simple form, so variants \appdu, \appdo and \apppo are identical. For \emph{cubes} and
\emph{opt} all nonlinear
terms involve a single multiplication or exponentiation, so for these experiments also variants
\apppu and \appfe are the same.
For the \emph{fractions} problem, and for \emph{sumprod},
no exponentiations are used, so all
three variants of the full decomposition
approach that we implemented are identical.

The results of these experiments clearly show
the disadvantage of implementing exponentiation by means of multiplication:
there is less domain reduction
because we increase the number of variable occurrences
(see the dependency problem, discussed in Section~\ref{sec:box}).
For \emph{opt} and variant \appfm, the run did not complete within reasonable
time and was aborted.

For \emph{fractions} the symbolic manipulation
of variant \appdo reduces the search tree by a factor $0.70$.
However, this reduction
is not reflected in the timings, and in fact the CPU time even increases.
The reason is that computing the domain updates involves adding
intervals of real numbers.
The arithmetic operations on such intervals are more expensive than
their counterparts on integer intervals, because the bounds have to be
maintained as rational numbers.
Arithmetic operations on rational numbers are more expensive because they
involve the computation of greatest common divisors.
For \emph{kyoto} the symbolic manipulation did not reduce the size of the
search tree, so the effect is even more severe.

In general, the introduction of auxiliary variables leads to
a reduction of the number of interval operations compared to
the direct approach.
As discussed at the end of Section~\ref{sec:second}, this is because
auxiliary variables prevent the evaluation of subexpressions
that did not change. This effect is strongest for \emph{fractions}, where
the main constraint contains a large number of different power products.
Without auxiliary variables all power products are evaluated for every
\emph{POLYNOMIAL EQUALITY} rule defined by this constraint, even those
power products the variable domains of which did not change. With auxiliary variables
the intervals for such unmodified terms are available immediately, which leads
to a significant reduction of the number of interval multiplications.
For \emph{sumprod}, the difference between variants \appd and \apppu is a bit
artificial, because the operations that are saved involve the computation
of the constant term $c_1\cdot\ldots\cdot c_n$. A~comparable number
of interval additions can be saved if we introduce a variable for
the constant term $c_1+\ldots+c_n$. If we add these variables to the CSP
all variants of
the direct and partial decomposition approaches
are essentially the same.

That stronger reduction is achieved as a result of
full decomposition,
mentioned in Section~\ref{sec:third}, is seen for
the \emph{fractions} benchmark and more prominently
for \emph{sumprod}. In the latter benchmark, this effect depends
on a decomposition of the term $\prod_{i=1}^{n}x_i$ as
$x_1\cdot(x_2\cdot(\ldots\cdot(x_{n-1}\cdot x_n)\ldots))$,
with an auxiliary variable for each pair of matching brackets.
The decomposition then
matches the chronological ordering used to select the variable for
branching. If the ordering is reversed, the number of nodes is equal
to that of the other approaches. The effect described
in Section~\ref{sec:second} is not demonstrated by these experiments.

If we do not consider the symbolic manipulation of variant \appdo,
variant \appfe leads to the smallest total number of interval
operations in all cases, but the scheduling mechanism discussed in
Section~\ref{sec-experiments} is essential for a consistent good performance.
If for example the schedule is omitted for \emph{opt}, the number of interval
operations almost triples, and performance of variants \apppu and \appfe
is then much worse than that of \appdu.
This is conform the observations of \cite{sac05goualard}, where it is
demonstrated that for constraints on reals, enforcing hull consistency
for a decomposition through repeated application of the HC4revise
procedure yields superior performance compared to the basic HC3 algorithm.
Based on these observations, we expect that the benefit of using the
schedule will grow with the number of variables.

The total numbers of interval operations in table~\ref{tab-numbers} do not
fully explain all differences in elapsed times.
One of the reasons is that different interval operations have different costs.
Also some overhead is involved in applying a reduction rule, so if the number of
applications differs significantly for two experiments, this
influences the elapsed times as well (\emph{opt\/}, \appd, \apppu).
The elapsed times are not the only measure
that is subject to implementation details. For example, we implemented division
by a constant interval $[-1..-1]$ as multiplication by a constant, which
is more efficient in our implementation. Such decisions are
reflected in the numbers reported in table~\ref{tab-numbers}.

For each of the benchmarks,
the last column of Table~\ref{tab-elapsed} compares the performance of
the variants that we implemented
with that of an \eclipse \cite{eclipsetutorial} program
that directly encodes the
problem statement of Subsection~\ref{subsec:problems} using the ic library.
For each problem, the first entry in this column lists the CPU time
reported by \eclipse for an all-solution search, where we applied
the same branching scheme as we used in OpenSolver. The other entries compare
propagation strength, for which we ran the solvers without search, and
compared the resulting domains of the variables.
A mark `=' means that the computed domains are the same, `+' that our
variant
achieved stronger reduction, and `-' that constraint propagation is weaker
than with \eclipse.

In addition, for \emph{cubes} we verified that the
number of nodes in the \eclipse search tree is identical to that for
all variants except \appfm and \appfs, which nicely fits with the
comparable CPU times.
In contrast, for the \emph{kyoto} benchmark, the number of
nodes in the search tree is substantially lower for our approaches than for
\eclipse, and so is the CPU time.
For the \emph{opt} problem the CPU time for our approaches (except \appfm) is also
substantially lower than for \eclipse. We have not verified the number of
nodes visited by the \texttt{minimize/2} built-in, but the sequence of
suboptimal solutions is identical to that found by our approaches
(not verified for \appfm).
For this comparison we used \eclipse version 5.10.

\section{Conclusions}
\label{sec:conclusions}
In this paper we discussed a number of approaches to constraint
propagation for arithmetic constraints on integer intervals. To assess
them we implemented them using
the OpenSolver framework of \cite{pz03os},
and compared their
performance on a number of benchmark problems. We can conclude that:
\begin{itemize}
   \item
      Implementation of exponentiation by multiplication gives weak reduction.
      In
      the full decomposition approach
      $x = y^n$ should be used as an
      atomic constraint.
   \item
      The optimization of the direct approach, where common powers
      of variables are divided out, can significantly reduce the
      size of the search tree, but the resulting reduction steps
      rely heavily on the division and addition of rational numbers.
      These operations are more expensive than their
      integer counterparts, because they involve the computation of
      greatest common divisors.
      As a result, our implementation of this approach was inefficient.

   \item Introducing auxiliary variables can be beneficial in two
      ways: it may strength\-en constraint propagation, as discussed in
      Sections~\ref{sec:second} and~\ref{sec:third},
      and it may prevent the evaluation of
      subexpressions the variable domains of which did not change.

   \item
      As a result, given an appropriate scheduling of the rules,
      the full and partial decomposition approaches perform better
      than the direct approach without the optimization,
      in terms of numbers of interval operations. Actual performance
      depends on many implementation aspects. However for our test problems
      the performance of variants \apppu, \apppo and \appfe does not differ much,
      except for one case
      where the decomposition of a single multiplication of all variables
      significantly reduced the size of the search tree.
\end{itemize}

Because of the inherent simplicity of the reduction rules and the potential
reduction of the search tree, full decomposition of arithmetic constraints
into multiplication, exponentiation, and linear constraints is our method of
choice.
However, a hierarchical scheduling of the resulting
reduction rules is essential for efficient constraint propagation, and
if a solver does not provide facilities for controlling the propagation order,
the direct approach is preferable.

Given that the optimization of the direct approach
can achieve a significant reduction of the search tree, 
it would be interesting to combine it with full decomposition. Depending on the
effect of the symbolic manipulation, a selection of the optimized
rules that enforce a particular constraint according to
variant \appdo
could be used as redundant rules.
In this case, the internal computations need not be precise,
and we could maintain the rational bounds as floating-point numbers,
thus avoiding the expensive computation of greatest common divisors.

\subsection*{Acknowledgements}
We would like to thank Maarten van Emden and the referees of earlier versions 
for useful comments.

\bibliographystyle{plain}

\begin{thebibliography}{10}

\bibitem{Apt98a}
K.~R. Apt.
\newblock A proof theoretic view of constraint programming.
\newblock {\em Fundamenta Informaticae}, 33(3):263--293, 1998.
\newblock Available via \url{http://arXiv.org/archive/cs/}.

\bibitem{Apt03}
K.~R. Apt.
\newblock {\em Principles of Constraint Programming}.
\newblock Cambridge University Press, 2003.

\bibitem{AZ04}
K.~R. Apt and P.~Zoeteweij.
\newblock A comparative study of arithmetic constraints on integer intervals.
\newblock In K.~R. Apt, F.~Fages, F.~Rossi, P.~Szeredi, and J.~V\'ancza,
  editors, {\em Recent Advances in Constraints}, volume 3010 of {\em Lecture
  Notes in Artificial Intelligence}, pages 1--24, 2004.
\newblock Available via \url{http://xxx.lanl.gov/abs/cs.DS/0403016}.

\bibitem{granvilliers99revising}
F.~Benhamou, F.~Goualard, L.~Granvilliers, and J.-F. Puget.
\newblock Revising hull and box consistency.
\newblock In {\em Proceedings of the 16th International Conference on Logic
  Programming ({ICLP}'99)}, pages 230--244. The MIT Press, 1999.

\bibitem{benhamou94revisited}
F.~Benhamou, D.~McAllester, and P.~Van~Hentenryck.
\newblock {CLP(intervals) revisited}.
\newblock In {\em Proceedings of the 1994 International Symposium on Logic
  programming}, pages 124--138. MIT Press, 1994.

\bibitem{BenOld97}
F.~Benhamou and W.~J. Older.
\newblock Applying interval arithmetic to real, integer, and {Boolean}
  constraints.
\newblock {\em Journal of Logic Programming}, 32(1):1--24, 1997.

\bibitem{eclipsetutorial}
A.~M. Cheadle, W.~Harvey, A.~J. Sadler, J.~Schimpf, K.~Shen, and M.~G. Wallace.
\newblock {\eclipse: a Tutorial Introduction}.
\newblock Available from \url{http://www.eclipse-clp.org/}.

\bibitem{clpfd}
P.~Codognet and D.~Diaz.
\newblock Compiling constraints in {\texttt{clp(fd)}}.
\newblock {\em Journal of Logic Programming}, 27(3):185--226, 1996.

\bibitem{collaviza99comparing}
H.~Collaviza, F.~Delobel, and M.~Rueher.
\newblock Comparing partial consistencies.
\newblock {\em Reliable Computing}, 5(3):213--228, 1999.

\bibitem{cyberteaser}
V.~Dubrovsky and A.~Shvetsov.
\newblock {\textit{Quantum}} {Cyberteaser}.
\newblock Available from \url{http://www.nsta.org/quantum/kyotoarc.asp},
  {May/June} 1995.

\bibitem{gmp}
{\em {GNU MP: The GNU Multiple Precision Library}}, September 2006.
\newblock Edition 4.2.1, available from \url{http://www.gmplib.org}.

\bibitem{sac05goualard}
F.~Goualard and L.~Granvilliers.
\newblock Controlled propagation in continuous numerical constraint networks.
\newblock In {\em Proceedings of the 2005 ACM symposium on Applied computing},
  pages 377--382. ACM Press, 2005.

\bibitem{realpaver}
L.~Granvilliers.
\newblock {\em {RealPaver} User's Manual: Solving Nonlinear Constraints by
  Interval Computations}, August 2004.
\newblock Edition 0.4, for RealPaver Version 0.4, available from
  \url{http://sourceforge.net/projects/realpaver}.

\bibitem{HS03}
W.~Harvey and P.~J. Stuckey.
\newblock Improving linear constraint propagation by changing constraint
  representation.
\newblock {\em Constraints}, 8(2):173--207, 2003.

\bibitem{HJvE01}
T.~J. Hickey, Q.~Ju, and M.~H. {van Emden}.
\newblock Interval arithmetic: from principles to implementation.
\newblock {\em Journal of the ACM}, 48(5):1038--1068, 2001.

\bibitem{MS98}
K.~Marriott and P.~Stuckey.
\newblock {\em Programming with Constraints: An Introduction}.
\newblock The MIT Press, 1998.

\bibitem{Moo66}
R.~E. Moore.
\newblock {\em Interval Analysis}.
\newblock Prentice-Hall, Englewood Cliffs, NJ, 1966.

\bibitem{Rat96}
D.~Ratz.
\newblock Inclusion isotone extended interval arithmetic: a toolbox update.
\newblock Technical Report D-76128, University of Karlsruhe, 1996.

\bibitem{SS02}
C.~Schulte and G.~Smolka.
\newblock Finite domain constraint programming in {Oz}. {A} tutorial.
\newblock Available from \url{http://www.mozart-oz.org/documentation/fdt/},
  August 2002.
\newblock Version 1.3.1 (20040616).

\bibitem{SS05}
C.~Schulte and P.~J. Stuckey.
\newblock When do bounds and domain propagation lead to the same search space?
\newblock {\em Transactions on Programming Languages and Systems},
  27(3):388--425, May 2005.

\bibitem{numerica}
P.~Van~Hentenryck, L.~Michel, and Y.~Deville.
\newblock {\em Numerica: a Modeling Language for Global Optimization}.
\newblock The MIT Press, 1997.

\bibitem{cai92wallace}
R.~J. Wallace and E.~F. Freuder.
\newblock Ordering heuristics for arc consistency algorithms.
\newblock Available from \url{http://4c.ucc.ie/web/pubs.jsp}, 1992.

\bibitem{pz03os}
P.~Zoeteweij.
\newblock Opensolver: A coordination-enabled abstract branch-and-prune tree
  search engine (abstract).
\newblock In F.~Rossi, editor, {\em Proceedings of CP 2003}, volume 2833 of
  {\em LNCS}, page 1002. Springer-Verlag, 2003.

\end{thebibliography}

\section*{Appendix}

We provide here the proofs of the
Bounds consistency Theorems \ref{thm:multi} and \ref{thm:multi2}, and the
\emph{MULTIPLICATION} Theorem \ref{thm:multi3}.
\medskip

\NI
\textbf{Proof of the Bounds consistency Theorem \ref{thm:multi}}.

\NI
Let $\phi := \p{x \cdot y = z}{x \in D_x, y \in D_y,z \in D_z}$.
Call a variable $u$ of $\phi$ \emph{bounds consistent} if the bounds of its
domain satisfy the condition of the bounds consistency
(see Definition \ref{def:bounds}).

Given an integer interval $[l..h]$ denote by $\overline{[l..h]}$
the corresponding real interval $[l, h]$.
Suppose that $D_x = [l_x .. h_x], D_y = [l_y .. h_y], D_z = [l_z .. h_z]$.  
To show that $\phi$ is closed under the applications of
the \emph{MULTIPLICATION 1} rule it suffices to prove that
\begin{equation}
  \label{eq:dz}
\C{l_z, h_z}\sse \opint(D_x \cdot D_y).  
\end{equation}

So take $c \in \C{l_z, h_z}$.  By the bounds consistency of $z$ we have $c = a \cdot b$ for some $a
\in \overline{D_x}$ and $b \in \overline{D_y}$.  Since $D_x$ and $D_y$
are integer intervals we have $\floor{a}, \ceiling{a} \in D_x$ and
$\floor{b}, \ceiling{b} \in D_y$.  To prove (\ref{eq:dz}), by the
definition of $D_x \cdot D_y$, we need to find $a_1, a_2 \in D_x$ and
$b_1, b_2 \in D_y$ such that 
\[
a_1 \cdot b_1 \leq c \leq a_2 \cdot b_2.
\]
The choice of $a_1, a_2, b_1$ and $b_2$ depends on the sign of
$a$ and of $b$ and is provided in the following table:
\medskip

\begin{center}
\begin{tabular}{|l|l|l|l|l|} \hline
condition    & $a_1$               &  $b_1$        &  $a_2$        &  $b_2$ \\ \hline
$a = 0$      & $a$                 & $\floor{b}$   & $a$           & $\floor{b}$  \\
$b = 0$      & $\floor{a}$         & $b$           & $\floor{a}$   & $b$ \\
$a >0, b >0$ & $\floor{a}$         & $\floor{b}$   & $\ceiling{a}$ & $\ceiling{b}$ \\
$a >0, b <0$ & $\ceiling{a}$       & $\floor{b}$   & $\floor{a}$   & $\ceiling{b}$ \\
$a <0, b >0$ & $\floor{a}$         & $\ceiling{b}$ & $\ceiling{a}$ & $\floor{b}$ \\
$a <0, b <0$ & $\ceiling{a}$       & $\ceiling{b}$ & $\floor{a}$   & $\floor{b}$   \\
\hline
\end{tabular}
\end{center}
\medskip

To prove 
that $\phi$ is closed under the applications of the
\emph{MULTIPLICATION 2} and \emph{3} rules 
it suffices to prove

\begin{equation}
  \label{eq:intervals}
\mbox{$\C{l_x, h_x} \sse \opint(D_z / D_y)$ and $\C{l_y, h_y} \sse \opint(D_z / D_x)$}.
\end{equation}

We need to distinguish a number of cases. The case analysis depends on the 
position of 0 w.r.t.~each of the intervals $D_x$ and $D_y$. This leads to 9 cases, which 
by symmetry between $x$ and $y$ can be reduced to 6 cases. We present here the proofs
for representative 3 cases.

\medskip

\NI
\emph{Case 1}. $l_x \geq 0, \ l_y \geq 0$.

By the bounds consistency of $x$ for some $b \in [l_y, h_y]$ we have
$l_x \cdot b \in [l_z, h_z]$. Then $b \leq h_y$ and $l_x \geq 0$, so
$l_x \cdot b \leq l_x \cdot h_y$.
Also $l_z \leq l_x \cdot b$, so
\[
l_z \leq l_x \cdot h_y.
\]

Next, by the bounds consistency of $y$ for some $a \in [l_x, h_x]$ we
have $a \cdot h_y \in [l_z, h_z]$.  Then $l_x \leq a$ and $h_y \geq 0$,
so $l_x \cdot h_y \leq a \cdot h_y$. Also $a \cdot h_y \leq h_z$, so
\[
l_x \cdot h_y \leq h_z.
\]
So $l_x \cdot h_y \in [l_z .. h_z]$ and consequently
by the definition of the integer intervals division
\[
\mbox{$l_x \in D_z/D_y$ and $h_y \in D_z/D_x$}.
\]
By a symmetric argument
\[
\mbox{$h_x \in D_z/D_y$ and $l_y \in D_z/D_x$}.
\]
\medskip

\NI
\emph{Case 2}. $l_x \geq 0, \ h_y \leq 0$.

By the bounds consistency of $x$ for some $b \in [l_y, h_y]$ we have
$h_x \cdot b \in [l_z, h_z]$. Then $b \leq h_y$ and $h_x \geq 0$, so
$h_x \cdot b \leq h_x \cdot h_y$.
Also $l_z \leq h_x \cdot b$, so
\[
l_z \leq h_x \cdot h_y.
\]

Next, by the bounds consistency of $y$ for some $a \in [l_x, h_x]$ we
have $a \cdot h_y \in [l_z, h_z]$.  Then $a \leq h_x$ and $h_y \leq 0$,
so $a \cdot h_y \geq h_x \cdot h_y$. Also $h_z \geq a \cdot h_y$, so
\[
h_x \cdot h_y \leq h_z.
\]
So $h_x \cdot h_y \in [l_z .. h_z]$ and consequently
by the definition of the integer intervals division
\[
\mbox{$h_x \in D_z/D_y$ and $h_y \in D_z/D_x$}.
\]

Further, by the bounds consistency of $x$ for some $b \in [l_y, h_y]$ we have
$l_x \cdot b \in [l_z, h_z]$. Then $l_y \leq b$ and $l_x \geq 0$, so
$l_x \cdot l_y \leq l_x \cdot b$. Also $l_x \cdot b \leq h_z$, so
\[
l_x \cdot l_y \leq h_z.
\]
Next, by the bounds consistency of $y$ for some $a \in [l_x, h_x]$ we
have $a \cdot l_y \in [l_z, h_z]$.  Then $l_x \leq a$ and $l_y < 0$,
so $l_x \cdot l_y \geq a \cdot l_y$. 
Also $a \cdot l_y \geq l_z$, so
\[
l_z \leq l_x \cdot l_y.
\]
So $l_x \cdot l_y \in [l_z .. h_z]$ and consequently
by the definition of the integer intervals division
\[
\mbox{$l_x \in D_z/D_y$ and $l_y \in D_z/D_x$}.
\]
\medskip

\NI
\emph{Case 3}. $l_x < 0 < h_x, \ l_y \geq 0$.

The proof for this case is somewhat more elaborate.
By the bounds consistency of $x$ for some $b \in [l_y, h_y]$ we have
$l_x \cdot b \in [l_z, h_z]$.  Then $l_y \leq b$ and $l_x < 0$, so
$l_x \cdot l_y \geq l_x \cdot b$.  But also $l_x \cdot b \geq l_z$, so
\[
l_z \leq l_x \cdot l_y.
\]

Next, by the bounds consistency of $y$ for some $a \in [l_x, h_x]$ we
have $a \cdot l_y \in [l_z, h_z]$.  Then $l_x \leq a$ and $l_y \geq 0$, so $l_x \cdot l_y \leq a \cdot l_y$.
But also $a \cdot l_y \leq h_z$, so 
\[
l_x \cdot l_y \leq h_z.
\]
So  $l_x \cdot l_y \in [l_z .. h_z]$ and consequently
by the definition of the integer intervals division
\[
\mbox{$l_x \in D_z/D_y$ and $l_y \in D_z/D_x$}.
\]

Further, by the bounds consistency of 
$x$ for some $b \in [l_y, h_y]$ we have
$h_x \cdot b \in [l_z, h_z]$.  Then $l_y \leq b$ and $h_x > 0$, so $h_x \cdot l_y  \leq h_x \cdot b$.
But also $h_x \cdot b \leq h_z$, so 
\[
h_x \cdot l_y  \leq h_z.
\]

Next, we already noted that by the bounds consistency of 
$y$ for some $a \in [l_x, h_x]$ we
have $a \cdot l_y \in [l_z, h_z]$. Then
$a \leq h_x$ and $l_y \geq 0$, so $a \cdot l_y \leq h_x \cdot l_y$.
But also $l_z \leq a \cdot l_y$, so 
\[
l_z \leq h_x \cdot l_y.
\]
So $h_x \cdot l_y \in [l_z .. h_z]$ and consequently
by the definition of the integer intervals division
\[
\mbox{$h_x \in D_z/D_y$}.
\]

It remains to prove that $h_y \in D_z/D_x$.
We showed already $l_x \cdot l_y \leq h_z$.
Moreover, $l_x < 0$ and $l_y \leq h_y$, so 
$l_x \cdot h_y \leq l_x \cdot l_y$ and hence
\[
l_x \cdot h_y \leq h_z.
\]

Also we showed already $l_z \leq h_x \cdot l_y$. Moreover $h_x > 0$
and $l_y \leq h_y$, so  $h_x \cdot l_y \leq h_x \cdot h_y$ and hence
\[
l_z \leq h_x \cdot h_y.
\]
So if either $l_z \leq l_x \cdot h_y$ or $h_x \cdot h_y \leq h_z$, then
either $l_x \cdot h_y \in [l_z .. h_z]$ or $h_x \cdot h_y \in [l_z .. h_z]$
and consequently $h_y \in D_z/D_x$.

If both $l_x \cdot h_y < l_z$ and $h_z < h_x \cdot h_y$, then
\[
[l_z .. h_z] \sse [l_x .. h_x] \cdot h_y.  
\]
In particular
for some $a \in D_x$ we have $l_z = a \cdot h_y$, so 
$h_y \in D_z / D_x$, as well.

This concludes the proof for this case.
\HB
\bigskip

\NI
\textbf{Proof of the Bounds consistency Theorem \ref{thm:multi2}}.

\NI
We consider each variable in turn. We begin with
$x$. Suppose that $D_x = [l_x .. h_x]$. 
$\phi$ is closed
under the applications of the \emph{MULTIPLICATION 2} rule, so
\begin{equation}
  \label{eq:dx}
\C{l_x, h_x} \sse \opint(D_z/D_y).  
\end{equation}

To show the bounds consistency of $x$ amounts to showing
\begin{equation}
  \label{eq:included}
\C{l_x, h_x} \sse \overline{D_z} \oslash \overline{D_y}.
\end{equation}
(Recall that given real intervals $X$ and $Y$ we denote by $X \oslash Y$
their division, defined in Section \ref{sec:interval}.)
\medskip

\NI
\emph{Case 1}. $\opint(D_z/D_y) = {\cal Z}$. 

This implies that $0 \in
D_z \cap D_y$, so by the definition of real intervals division
$\overline{D_z} \oslash \overline{D_y} = \langle - \infty, \infty \rangle$.
Hence (\ref{eq:included}) holds.  
\medskip

\NI
\emph{Case 2}. $\opint(D_z/D_y) \neq {\cal Z}$. 

So $\opint(D_z/D_y)$ is an integer interval, say
$\opint(D_z/D_y) = [l_{zy} .. h_{zy}]$.
Two subcases arise.
\medskip

\NI
\emph{Subcase 1}. $\overline{D_z} \oslash \overline{D_y}$ is a, possibly open ended, real interval.

By (\ref{eq:dx}) for some $b_1, b_2 \in D_y$ and $c_1, c_2 \in D_z$
we have
\[
l_{zy} \cdot b_1 = c_1,
\]
\[
h_{zy} \cdot b_2 = c_2.
\]

Let 
\[
\underline{b} := \opmin(b_1, b_2), \overline{b}  := \opmax(b_1, b_2), \underline{c} := \opmin(c_1, c_2), \overline{c}  := \opmax(c_1, c_2).
\]

So $\C{l_{zy}, h_{zy}} \sse [\underline{c}, \overline{c}] \oslash [\underline{b}, \overline{b}]$. 
Also $[\underline{c}, \overline{c}] \oslash
[\underline{b}, \overline{b}] \sse \overline{D_z} \oslash \overline{D_y}$. Hence $\C{l_{zy}, h_{zy}}
\sse \overline{D_z} \oslash \overline{D_y}$ and consequently, by the
assumption for this subcase, $[l_{zy}, h_{zy}] \sse \overline{D_z} \oslash
\overline{D_y}$. This proves (\ref{eq:included}) since by (\ref{eq:dx}) 
$\C{l_x, h_x} \sse [l_{zy}, h_{zy}]$.
\medskip

\NI
\emph{Subcase 2}. $\overline{D_z} \oslash \overline{D_y}$ is not a, possibly open ended, real interval.

In what follows for an integer interval $D := [l .. h]$ we write $D
>0$ if $l > 0$, $D < 0$ if $h < 0$. Also recall that 
$\langle D \rangle := \C{x \in {\cal Z} \mid l < x < h}$.

This subcase can arise only when $D_z > 0$ and $0
\in \langle D_y \rangle$ or $D_z < 0$ and $0 \in \langle D_y \rangle$,
see \cite{Rat96} (reported as Theorem 4.8 in
\cite{HJvE01}), where the definition of the division of real intervals
is considered.

Since $\phi$ is closed under the \emph{MULTIPLICATION} rule 3
\[
D_y \sse \opint(D_z/D_x). 
\]
So $\opint(D_z/D_x) \neq \ES$ since by assumption $D_y$ is non-empty.
Also, since $0 \not\in D_z$, we have $\opint(D_z/D_x) \neq {\cal Z}$.
So $\opint(D_z/D_x)$ is a non-empty integer interval such that $0 \in \langle \opint(D_z/D_x) \rangle$.

But $D_z > 0$ or $D_z <0$, so if $D_x > 0$, then $\opint(D_z / D_x) > 0 $
or $\opint(D_z / D_x) < 0$ and if $D_x < 0$, then $\opint(D_z / D_x) > 0 $
or $\opint(D_z / D_x) < 0$, as well.  So $0 \in \langle D_x \rangle$.
Hence $0 \in \langle D_x \rangle \cap \langle D_y \rangle$ while $0 \not\in D_z$.  This contradicts
(\ref{eq:condition}). So this subcase cannot arise.

\medskip

The proof for the variable $y$ is symmetric to the one for the variable $x$.
\medskip

Consider now the variable $z$. 
$\phi$ is closed under the applications of
the \emph{MULTIPLICATION 1} rule, so
\[
D_z \sse \opint(D_x \cdot D_y).
\]

Take now $c \in D_z$.  Then there exist $a_1, a_2 \in D_x$ and $b_1, b_2
\in D_y$ such that $ a_1 \cdot b_1 \leq c \leq a_2 \cdot b_2.  $ We
can assume that both inequalities are strict, that is,
\begin{equation}
\label{eq:strict}
a_1 \cdot b_1 < c < a_2 \cdot b_2,
\end{equation}
since otherwise the desired conclusion is established.

Let
\[
\underline{a} := \opmin(a_1, a_2), \overline{a}  := \opmax(a_1, a_2), \underline{b} := \opmin(b_1, b_2), \overline{b}  := \opmax(b_1, b_2).
\]

We now show that $a \in [\underline{a}.. \overline{a}]$
and $b \in [\underline{b}.. \overline{b}]$ exist such that $c = a \cdot b$.  
Since $[\underline{a}.. \overline{a}] \sse \overline{D_x}$ and
$[\underline{b}.. \overline{b}] \sse \overline{D_y}$, this will establish
the bounds consistency of $z$.

The choice of $a$ and $b$ depends on the signs of $a_1$ and
$b_2$. When one of these values is zero, the choice is provided in the
following table, where in each case on the account of
(\ref{eq:strict}) no division by zero takes place: 
\medskip

\begin{center}
\begin{tabular}{|l|l|l|} \hline
condition      & $a$      &  $b$ \\ \hline
$a_1 = 0$      & $c/b_2$  &  $b_2$ \\
$a_2 = 0$      & $c/b_1$  &  $b_1$ \\
$b_1 = 0$      & $a_2$    &  $c/a_2$ \\
$b_2 = 0$      & $a_1$    &  $c/a_1$ \\
\hline
\end{tabular}
\end{center}
\medskip

It is straightforward to show that in each case the quotient belongs
to the corresponding interval.  For example, when $a_1 = 0$ we need to
prove that $c/b_2 \in [\underline{a}.. \overline{a}]$.  By
(\ref{eq:strict}) $a_2 \neq 0$. If $a_2 > 0$, then
 again by (\ref{eq:strict}), $b_2 > 0$, so
$c/b_2 \in [0.. a_2]$.  In turn, if $a_2 <0$, then also by
(\ref{eq:strict}) $b_2 < 0$, so, yet again by
(\ref{eq:strict}), $c/b_2 \in [a_2 .. 0]$.

When neither $a_1$ nor $b_2$ is zero, the choice
of $a$ and $b$ has to be argued case by case.
\medskip

\NI
\emph{Case 1}. $a_1 > 0, \ b_2 > 0$.

Then by (\ref{eq:strict}) $b_1 < c/a_1$ and $c/b_2 < a_2$.
Suppose that both $b_2 < c/a_1$ and $c/b_2 < a_1$.
Then 
$a_1 \cdot b_2 < c < a_1 \cdot b_2$, which is a contradiction.
So either $c/a_1 \leq b_2$ or $a_1 \leq c/b_2$, that is either
$c/a_1 \in [b_1 .. b_2]$ or $c/b_2 \in [a_1 ..a_2]$.
\medskip

\NI
\emph{Case 2}. $a_1 > 0, \ b_2 < 0$.

Then by (\ref{eq:strict}) $b_1 < c/a_1$ and $a_2 < c/b_2$.
Suppose that both $b_2 < c/a_1$ and $a_1 < c/b_2$.
Then $a_1 \cdot b_2 < c < a_1 \cdot b_2$, which is a contradiction.
So either $c/a_1 \leq b_2$ or $c/b_2 \leq a_2$, that is either
$c/a_1 \in [b_1 .. b_2]$ or $c/b_2 \in [a_2 ..a_1]$.
\medskip

\NI
\emph{Case 3}. $a_1 < 0, \ b_2 > 0$.

Then by (\ref{eq:strict}) $c/a_1 < b_1$ and $c/b_2 < a_2$.
Suppose that both $c/a_1 < b_2$ and $c/b_2 < a_1$.
Then $a_1 \cdot b_2 < c < a_1 \cdot b_2$, which is a contradiction.
So either $b_2 \leq c/a_1$ or $a_1 \leq c/b_2$, that is either
$c/a_1 \in [b_2 .. b_1]$ or $c/b_2 \in [a_1 ..a_2]$.
\medskip

\NI
\emph{Case 4}. $a_1 < 0, \ b_2 < 0$.

Then by (\ref{eq:strict}) $c/a_1 < b_1$ and $a_2 < c/b_2$.
Suppose that both $c/a_1 < b_2$ and $a_1 < c/b_2$.
Then $a_1 \cdot b_2 < c < a_1 \cdot b_2$, which is a contradiction.
So either $b_2 \leq c/a_1$ or $c/b_2 \leq a_1$, that is either
$c/a_1 \in [b_2 .. b_1]$ or $c/b_2 \in [a_2 ..a_1]$.
\medskip

So in each of the four cases we can choose either $a := a_1$ and $b
:=c/a_1$ or $a := c/b_2$ and $b := b_2$.
\HB

\medskip

\NI
\textbf{Proof of the \emph{MULTIPLICATION} Theorem \ref{thm:multi3}}.

\NI 
The weak interval division produces larger sets than the interval
division. As a result the \emph{MULTIPLICATION} rules 2w and 3w yield
a weaker reduction than the original \emph{MULTIPLICATION} rules 2 and
3.  So it suffices to prove that $\phi := \p{x \cdot y = z}{x \in D_x,
  y \in D_y,z \in D_z}$ is closed under the applications of the
\emph{MULTIPLICATION} 1, 2 and 3 rules assuming that it is closed
under the applications of the \emph{MULTIPLICATION} 1, 2w and 3w
rules.  Suppose that $D_x = [l_x .. h_x], D_y = [l_y .. h_y], D_z =
[l_z .. h_z]$.
The assumption implies
\begin{equation}
   \{l_x,h_x\} \subseteq \opint(D_z:D_y)
   \label{eq:closed2w}
\end{equation}
and
\begin{equation}
   \{l_y,h_y\} \subseteq \opint(D_z:D_x)
   \label{eq:closed3w}
\end{equation}

The proof is by contradiction. Assume that (\ref{eq:closed2w}) and
(\ref{eq:closed3w}) hold, while $\phi$ is not closed under application of
\emph{MULTIPLICATION} 2 and 3. Without loss of generality, suppose that
\emph{MULTIPLICATION} 2 is the rule that can make a further reduction.
This is the case iff
\[
\opint(D_z/D_y)\subset \opint(D_z : D_y).
\]
By definition, the
proper inclusion implies that $l_y\geq0$ or $h_y\leq0$. Assume $l_y\geq0$,
the case for $h_y\leq0$ is similar.
Let $l_y':=\opmax(1,l_y)$, and
let $A:=\{l_z/l_y', l_z/h_y, h_z/l_y', h_z/h_y\}$,
and $B:=\{l_z/l_x,l_z/h_x,h_z/l_x,h_z/h_x\}$.
A further implication of the proper inclusion is that
one or both of $l_y'$ and $h_y$ do not have a multiple in $D_z$: otherwise
$\opmin(A)$ and $\opmax(A)$ would be elements of $D_z/D_y$, and
we would have $\opint(D_z : D_y)=\opint(D_z/D_y)$.
The cases for $l_y'$ and $h_y$ can be seen
in isolation, and their proofs are similar, so here we only consider
the case that $l_y'$ does not have a multiple in $D_z$.
In what follows we can assume
$0 \notin D_z$, since otherwise $l_y'$ and $h_y$ do have a multiple in $D_z$.

\bigskip
\noindent
\emph{Case 1}. $l_z > 0$.\\
From (\ref{eq:closed2w}) it follows that
$h_x\leq\lfloor\opmax(A)\rfloor$, which
for the case $l_y',h_y,l_z,h_z>0$ that we consider here implies
$h_x\leq\lfloor h_z/l_y'\rfloor$.
Because $[l_z..h_z]$ does not contain a multiple of $l_y'$, we have
$\lfloor h_z/l_y' \rfloor = \lfloor l_z/l_y' \rfloor$, so
\[
   h_x\leq\lfloor l_z/l_y' \rfloor.
\]
A further consequence of (\ref{eq:closed2w})
is that $l_x,h_x>0$. From (\ref{eq:closed3w}) it follows that
$l_y'\geq\lceil \opmin(B)\rceil$, which for $l_x,l_z>0$ implies
\[
   l_y'\geq\lceil l_z/h_x \rceil \geq l_z/h_x \geq l_z/\lfloor h_z/l_y' \rfloor.
\]
Because $l_y'$ is no divisor of $l_z$, and both numbers are positive,
we have $\lfloor l_z / l_y' \rfloor < l_z/l_y'$, and consequently
$l_y' > l_z / (l_z/l_y')$, leading to $l_y' > l_y'$,
which is a contradiction.

\bigskip
\noindent
\emph{Case 2. $h_z < 0$}.\\
Similarly, because $l_y',h_y>0$ and $l_z,h_z<0$,
it follows from (\ref{eq:closed2w}) that
$l_x\geq \lceil\opmin(A)\rceil = \lceil l_z/l_y' \rceil$,
and $l_x,h_x<0$. Because $[l_z..l_h]$ does not contain a multiple of $l_y'$,
we have $\lceil l_z/l_y'\rceil = \lceil h_z/l_y'\rceil$, so
\[
   l_x\geq\lceil h_z/l_y'\rceil.
\]
We use this information in the following implication of (\ref{eq:closed3w}):
\[
   l_y' \geq \lceil \opmin(B)\rceil = \lceil h_z/l_x' \rceil \geq h_z/l_x'
\]
to get
$l_y' \geq h_z / \lceil h_z/l_y' \rceil$.
Because $|\lceil h_z/l_y' \rceil| < |h_z/l_y'|$, we have
$l_y' > h_z / ( h_z/l_y')$, leading to $l_y' > l_y'$, which is a contradiction.
\HB

\end{document}